\documentclass[letterpaper]{article}
\usepackage{aaai25} 
\usepackage{times} 
\usepackage{helvet} 
\usepackage{courier} 
\usepackage[hyphens]{url} 
\usepackage{graphicx} 
\usepackage{graphicx}  
\usepackage{natbib}  
\usepackage{caption}  
\usepackage{algorithm}
\usepackage{algorithmic}

\usepackage{newfloat}
\usepackage{listings}
\DeclareCaptionStyle{ruled}{labelfont=normalfont,labelsep=colon,strut=off} 
\floatstyle{ruled}
\newfloat{listing}{tb}{lst}{}
\floatname{listing}{Listing}
\usepackage{amsmath,amsfonts,bm}

\def\eqref#1{equation~\ref{#1}}

\def\1{\bm{1}}

\def\vdelta{{\bm{\delta}}}
\def\vmu{{\bm{\mu}}}
\def\vmu{{\bm{\mu}}}

\def\vmu{{\bm{\mu}}}

\def\vsigma{{\bm{\sigma}}}

\def\vg{{\bm{g}}}

\def\vq{{\bm{q}}}

\def\vu{{\bm{u}}}
\def\vv{{\bm{v}}}
\def\vw{{\bm{w}}}

\DeclareMathAlphabet{\mathsfit}{\encodingdefault}{\sfdefault}{m}{sl}
\SetMathAlphabet{\mathsfit}{bold}{\encodingdefault}{\sfdefault}{bx}{n}

\def\gA{{\mathcal{A}}}
\def\gB{{\mathcal{B}}}

\def\gG{{\mathcal{G}}}
\def\gH{{\mathcal{H}}}

\def\gL{{\mathcal{L}}}

\def\gS{{\mathcal{S}}}

\DeclareMathOperator{\sign}{sign}

\usepackage{amsmath}
\usepackage{amsthm}
\usepackage{booktabs}
\usepackage[capitalize,noabbrev]{cleveref}
\usepackage{comment}
\usepackage{graphbox}
\usepackage{makecell}
\usepackage{subcaption}
\usepackage{xcolor}
\usepackage{xspace}

\theoremstyle{plain}

\newtheorem{proposition*}{Proposition}

\theoremstyle{definition}
\newtheorem{definition}{Definition}
\newtheorem{definition*}{Definition}

\newtheorem{assumption*}{Assumption}
\theoremstyle{remark}

\newtheorem{remark*}{Remark}

\crefname{fig}{Fig.}{Fig.}
\newcommand{\bvg}{\Bar{\vg}}
\newcommand{\hvg}{\hat{\vg}}

\newcommand{\norm}[1]{\lVert{#1}\rVert}
\newcommand{\ip}[2]{\langle{#1},{#2}\rangle}
\newcommand{\set}[1]{\{{#1}\}}
\newcommand{\abbreviation}{two-Stage aTtack based on gRadIent sKEw\xspace}
\newcommand{\SKEW}{STRIKE\xspace}

\title{Exploit Gradient Skewness\\to Circumvent Byzantine Defenses for Federated Learning\footnote{This paper is accepted by AAAI'25}}
\author{
    Yuchen Liu\textsuperscript{\rm 12}\equalcontrib\thanks{Work done during an internship at Sony AI.},
    Chen Chen\textsuperscript{\rm 3}\equalcontrib,
    Lingjuan Lyu\textsuperscript{\rm 3}\footnote{Corresponding author.},
    Yaochu Jin\textsuperscript{\rm 4},
    Gang Chen\textsuperscript{\rm 12}
}
\affiliations{
    \textsuperscript{\rm 1}The State Key Laboratory of Blockchain and Data Security, Zhejiang University\\
    \textsuperscript{\rm 2}Hangzhou High-Tech Zone (Binjiang) Institute of Blockchain and Data Security\\
    \textsuperscript{\rm 3}Sony AI\\
    \textsuperscript{\rm 4}{Westlake University, China}\\

    yuchen.liu.a@gmail.com,
    \{ChenA.Chen, lingjuan.lv\}@sony.com,
    {jinyaochu@westlake.edu.cn},
    cg@zju.edu.cn,
}

\begin{document}

\maketitle

\begin{abstract}
Federated Learning (FL) is notorious for its vulnerability to Byzantine attacks.
Most current Byzantine defenses share a common inductive bias: among all the gradients, the densely distributed ones are more likely to be honest.
However, such a bias is a poison to Byzantine robustness due to a newly discovered phenomenon in this paper -- gradient skew.
We discover that a group of densely distributed honest gradients skew away from the optimal gradient (the average of honest gradients) due to heterogeneous data.
This gradient skew phenomenon allows Byzantine gradients to hide within the densely distributed skewed gradients.
As a result, Byzantine defenses are confused into believing that Byzantine gradients are honest.
Motivated by this observation, we propose a novel skew-aware attack called \SKEW:
first, we search for the skewed gradients;
then, we construct Byzantine gradients within the skewed gradients.
Experiments on three benchmark datasets validate the effectiveness of our attack.
\end{abstract}

%
\begin{links}
    \link{Code}{https://github.com/YuchenLiu-a/byzantine_skew}
\end{links}

Federated Learning (FL) \cite{mcmahan2017fl, li2020fedprox} emerged as a privacy-aware learning paradigm, in which data owners, i.e., clients, repeatedly use their private data to compute local gradients and upload them to a central server.
The central server collects the uploaded gradients from clients and aggregates these gradients to update the global model.
In this way, clients can collaborate to train a model without exposing their private data.

Unfortunately, FL is susceptible to Byzantine attacks due to its distributed nature \cite{blanchard2017krum,guerraoui2018bulyan}.
A malicious party can control a small subset of clients, i.e., Byzantine clients, to degrade the utility of the global model.
During the training phase, Byzantine clients can send arbitrary messages to the central server to bias the global model.
A wealth of defenses \cite{blanchard2017krum,pillutla2019geometric,shejwalkar2021dnc} have been proposed to defend against Byzantine attacks in FL. 
They aim to estimate the optimal gradient, i.e., the average of gradients from honest clients, in the presence of Byzantine clients.

Most existing defenses \cite{blanchard2017krum,shejwalkar2021dnc,karimireddy2022bucketing} share a common inductive bias: the densely distributed gradients are more likely to be honest.
Generally, they assign higher weights to the densely distributed gradients. 
Then they compute the global gradient and use it to update the global model.
As a result, the output global gradient of defenses is biased towards the densely distributed of gradients.
\begin{figure}[t]
\begin{center}
\includegraphics[width=0.8\linewidth]{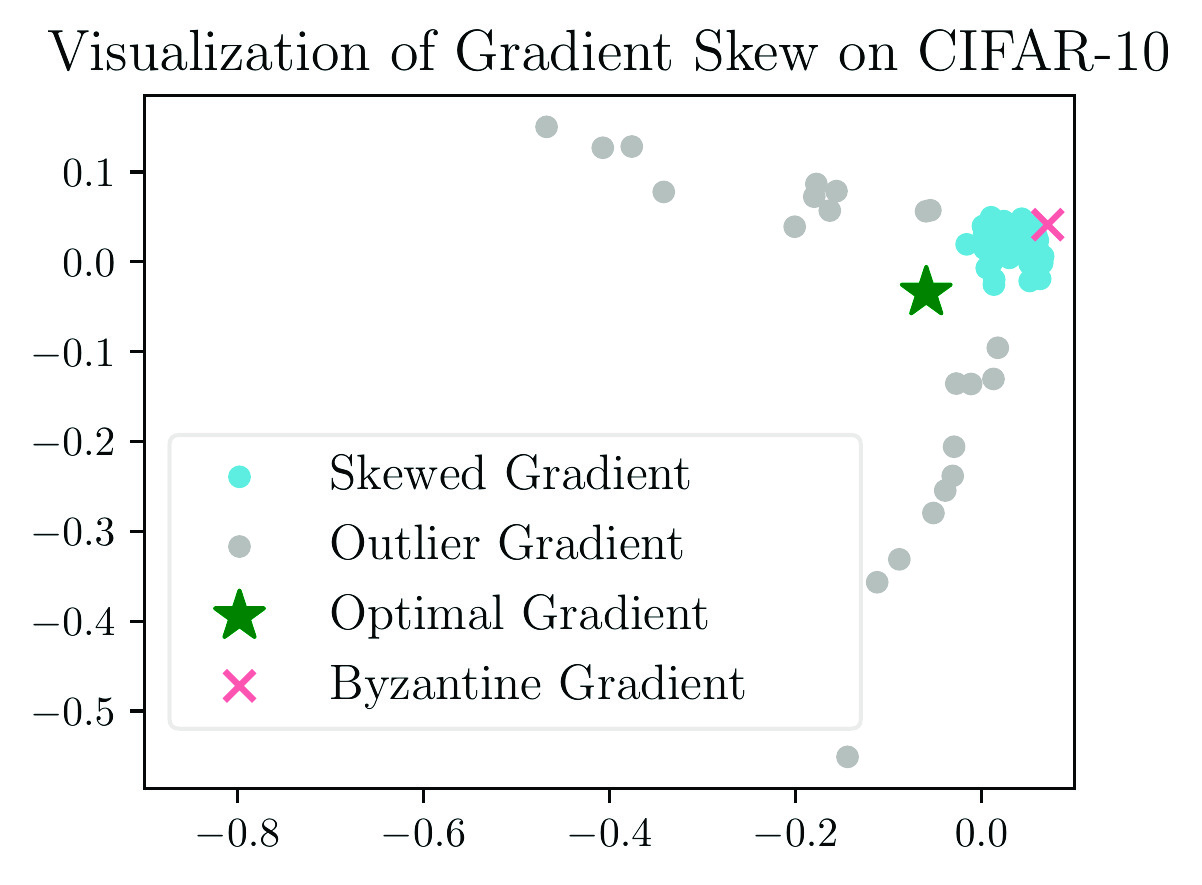}
\end{center}
\caption{
The LLE visualization of honest gradients in the non-IID setting on CIFAR-10.
Substantial honest gradients (blue circles) are skewed away from the optimal gradient (green star).
In this case, we can hide Byzantine gradients (pink crosses) within the skewed honest gradients to circumvent defenses.
}
\label{fig:main_visualization}
\end{figure}

However, this inductive bias of Byzantine defenses is harmful to Byzantine robustness in FL due to the presence of gradient skew.
In practical FL, data across different clients is non-independent and identically distributed (non-IID), which gives rise to heterogeneous honest gradients \cite{mcmahan2017fl,li2020fedprox,karimireddy2022bucketing}.
On closer inspection, we find that the distribution these heterogenous honest gradients are highly skewed.
In \cref{fig:main_visualization}, we use Locally Linear Embedding (LLE) \cite{roweis2000lle} to visualize the honest gradients on CIFAR-10 dataset \cite{krizhevsky2009cifar} when data is non-IID split.
Detailed setups and more results are provided in \cref{appsec:vis}.
As shown in \cref{fig:main_visualization}, a group of densely distributed gradients skews away from the optimal gradient.
We term this phenomenon as "gradient skew".
When honest gradients are skewed, the defenses' bias towards densely distributed gradients is a poison to Byzantine robustness.
In fact, we can hide Byzantine gradients within the skewed densely distributed honest gradients as shown in \cref{fig:main_visualization}.
In this case, the bias of defenses would drive the global gradient close to the skewed gradients but far from the optimal gradient.

In this paper, we
study how to exploit the gradient skew in the more practical non-IID setting to circumvent Byzantine defenses.
We first observe the gradient skew phenomenon in the non-IID setting and explore its vulnerability.
Motivated by the above observation, we design a novel \abbreviation called \SKEW.
In particular, \SKEW hides Byzantine gradients within the skewed honest gradients as shown in \cref{fig:main_visualization}.
\SKEW can take advantage of the gradient skew in FL to break Byzantine defenses.

In summary, our contributions are:
\begin{itemize}
    \item To the best of our knowledge, we are the first to discover the gradient skew phenomenon in FL:
    a group of densely distributed gradients is skewed away from the optimal gradient.
    Motivated by the observation, we design an attack principle that can circumvent Byzantine defenses under gradient skew:
    hide Byzantine gradients within the skewed honest gradients.
    \item Based on the above attack principle, we propose a two-stage Byzantine attack called \SKEW.
    In the first stage, \SKEW searches for the skewed honest gradients under the guidance of Karl Pearson's formula.
    In the second stage, \SKEW constructs the Byzantine gradients within the skewed honest gradients by solving a constrained optimization problem.
    \item Experiments on three benchmark datasets validate the effectiveness of the proposed attack.
    For instance, \SKEW attack improves upon the best baseline by 57.84\% against DnC on FEMNIST dataset when there are 20\% Byzantine clients.
\end{itemize}

\section{Related Works}

\textbf{Byzantine attacks.}
\citeauthor{blanchard2017krum} first disclose the Byzantine vulnerability of FL.
\citeauthor{baruch2019lie} observe that the variance of honest gradients is high enough for Byzantine clients to compromise Byzantine defenses. 
Based on this observation, they propose a LIE attack that hides Byzantine gradients within the variance.
\citeauthor{xie2020ipm} further utilize the high variance and propose an IPM attack.
Particularly, they show that when the variance of honest gradients is large enough, IPM can make the inner product between the aggregated gradient and the honest average negative.
However, this result is restricted to a few defenses, i.e., Median \cite{yin2018mediantrmean}, Trmean \cite{yin2018mediantrmean}, and Krum \cite{blanchard2017krum}.
\citeauthor{fang2020fang} establish an omniscient attack called Fang.
However, the Fang attack requires knowledge of the Byzantine defense, which is unrealistic in practice.
\citeauthor{shejwalkar2021dnc} propose Min-Max and Min-Sum attacks that solve a constrained optimization problem to determine Byzantine gradients.
From a high level, both Min-Max and Min-Sum aim to maximize the perturbation to a reference benign gradient while ensuring the Byzantine gradients lie within the variance.
\citeauthor{karimireddy2022bucketing} propose a Mimic attack that takes advantage of data heterogeneity in FL.
In particular, Byzantine clients pick an honest client to mimic and copy its gradient.
The above attacks take advantage of the large variance of honest gradients to break Byzantine defenses.
However, they all ignore the skewed nature of honest gradients in FL and fail to exploit this vulnerability.

\textbf{Byzantine resilience.}
\citeauthor{el2021collaborative, karimireddy2022bucketing} provide state-of-the-art theoretical analysis of Byzantine resilience under data heterogeneity.
\citeauthor{el2021collaborative} discuss Byzantine resilience in a decentralized, asynchronous setting.
\citeauthor{farhadkhani2022reasm} provide a unified framework for Byzantine resilience analysis, which enables comparison among different defenses on a common theoretical ground.
\citeauthor{karimireddy2022bucketing} improve the error bound of Byzantine resilience to be upper-bounded by the fraction of Byzantine clients, which recovers the standard convergence rate when there are no Byzantine clients.
\citeauthor{allouah2024byzantinerobust} tightly analyzing the impact of client subsampling and local steps.
\citeauthor{yan2024recess} utilizes the correlation of clients’ performance over multiple iterations to evaluate the reliability of clients.
They all share a common bias: densely distributed gradients are more likely to be honest.
However, this bias is a poison to Byzantine robustness in the presence of gradient skew.
In practical FL, the distribution of honest gradients is highly skewed due to data heterogeneity.
Therefore, existing defenses are especially vulnerable to attacks that are aware of gradient skew.

\textbf{Data heterogeneity.}
\citeauthor{yu2018parallel} first proposed to measure data heterogeneity by gradient divergence, which describes the difference between the local gradients and the global one. \citeauthor{karimireddy2020scaffold} proposed a more general version of gradient divergence - gradient dissimilarity. To the best of our knowledge, these are the only metrics of heterogeneity from a gradient distribution perspective \cite{li2019convergence,woodworth2020minibatch}. \citeauthor{luo2021no} find that such difference mainly involves neural network prediction heads. For label skewness, a particular type of heterogeneity, label distribution discrepancy is used to measure heterogeneity \cite{peng2024fedcal}. However, no existing work noticed that such gradient divergence is skewed - a group of densely distributed local gradients skew away from the global gradient, i.e., the gradient skew introduced in Section 4.

\section{Notations and Preliminary}
\subsection{Notations}
$\norm{\cdot}$ denotes the $\ell_2$ norm of a vector.
For vector $\vv$, $(\vv)_k$ represents the $k$-th coordinate of $\vv$.
Model parameters are denoted by $\vw$ and gradients are denoted by $\vg$.
We use $\bvg$ to denote the optimal gradient, i.e., the average of honest gradients, and $\hvg$ denotes the global gradients obtained by Byzantine defenses.
We use subscript $i$ to denote client $i$ and use superscript $t$ to denote communication round $t$.

\subsection{Preliminary}

\textbf{Federated learning.}
Suppose that there are $n$ clients and a central server.
The goal is to optimize the global loss function $\gL(\cdot)$:
\begin{align}
\min_\vw{\gL(\vw)},\quad\text{where }\gL(\vw)=\frac{1}{n}\sum_{i=1}^n\gL_i(\vw).
\end{align}
Here $\vw$ is the model parameter,
and $\gL_i(\cdot)$ is the local loss function on client $i$ for $i=1,\ldots,n$.

In communication round $t$, the central server distributes global parameter $\vw^t$ to the clients.
Each client $i$ performs several epochs of SGD to minimize its local loss function $\gL_i(\cdot)$ and update its local parameter to $\vw_i^{t+1}$.
Then, each client $i$ computes its local gradient $\vg_i^t$ and sends it to the server.
\begin{align}
\vg_i^t=\vw_i^t-\vw_i^{t+1},
\quad i=1,\ldots,n.
\end{align}
After receiving the uploaded local gradients, the server aggregates the local gradients and updates the global model to $\vw^{t+1}$.
\begin{gather}
\label{eq:mean}
\bvg^t=\frac{1}{n}\sum_{i=1}^n\vg_i^t,
\quad\vw^{t+1}=\vw^t-\bvg^t.
\end{gather}

\begin{figure*}[t]
\centering
\includegraphics[width=.75\linewidth]{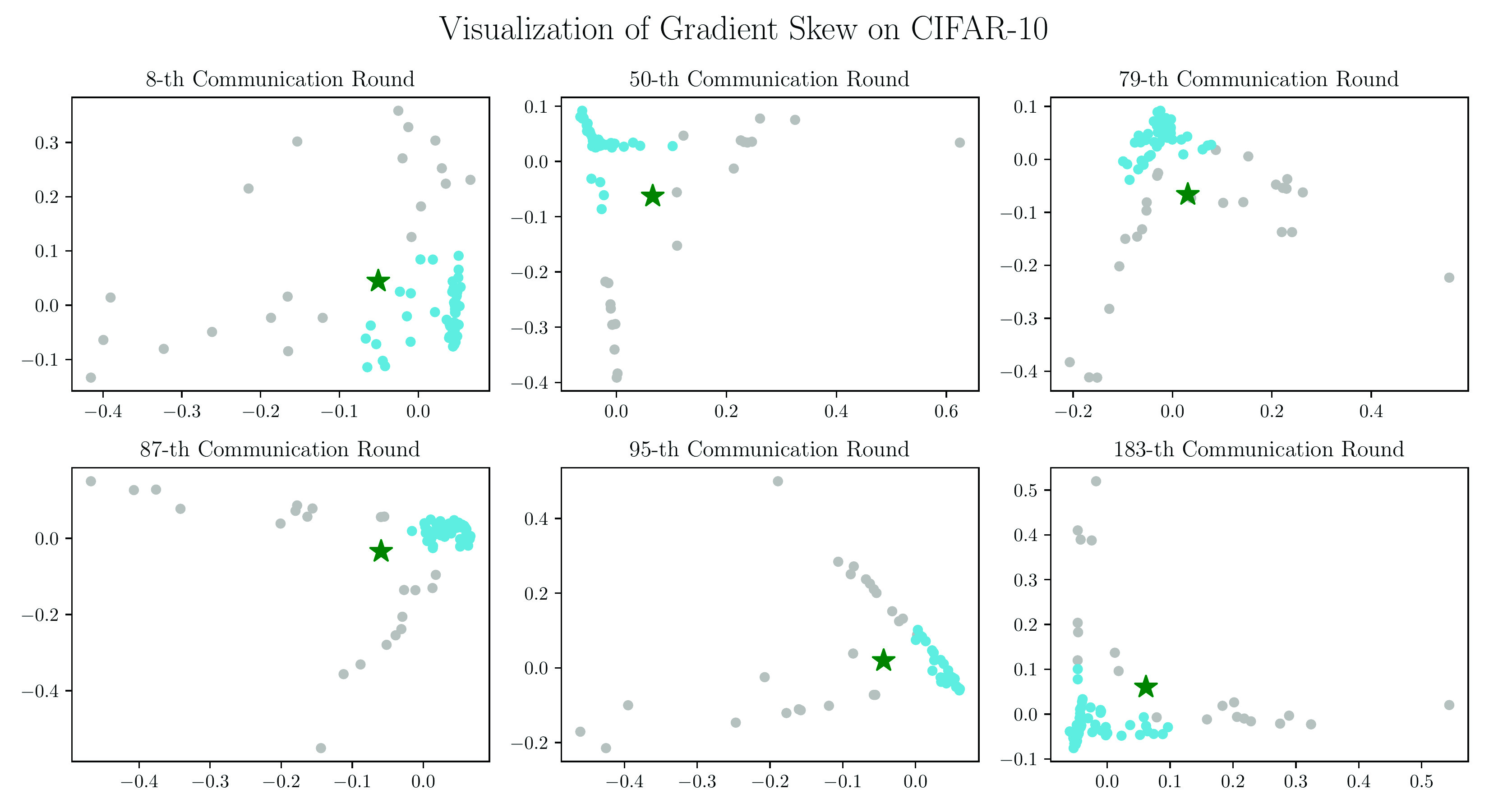}     
\includegraphics[width=.6\linewidth]{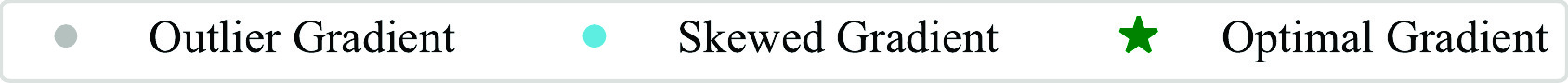}   \caption{Visualization of gradient skew on CIFAR-10 dataset. As shown in the figures, the optimal gradients (green stars) deviate from the densely distributed gradients.}
\label{fig:cifar10_vis}
\end{figure*}

\textbf{Byzantine attack model.}
Assume that among the total $n$ clients, $f$ fixed clients are Byzantine clients.
Let $\gB\subseteq\set{1,\ldots,n}$ denote the set of Byzantine clients and $\gH=\set{1,\ldots,n}\setminus\gB$ denote the set of honest clients.
In each communication round, Byzantine clients can send arbitrary messages to bias the global model.
The local gradients that the server receives in the $t$-th communication round are
\begin{align}
\vg_i^t=
\begin{cases}
*, & i\in\gB,\\
\vw^t-\vw_i^{t+1}, & i\in\gH,\\
\end{cases}
\end{align}
where $*$ represents an arbitrary message.
Following \cite{baruch2019lie,xie2020ipm}, we consider the setting where the attacker only has the knowledge of honest gradients.

\section{Gradient Skew in FL Due to Non-IID data}

\label{section:principle}
Plenty of works \cite{baruch2019lie,xie2020ipm,karimireddy2022bucketing} have explored how large variance can be harmful to Byzantine robustness.
However, to the best of our knowledge, none of the existing works is aware of the skewed nature of honest gradients in the non-IID setting and how gradient skew can threaten Byzantine robustness.

We take a close look at the distribution of honest gradients in the non-IID setting (without attack).
To construct our FL setup, we split CIFAR-10 \cite{krizhevsky2009cifar} dataset in a non-IID manner among 100 clients.
For more setup details, please refer to \cref{appsec:vis_setting}.
We run FedAvg \cite{mcmahan2017fl} for 200 communication rounds.
We randomly sample six communication rounds and use Locally Linear Embedding (LLE) \cite{roweis2000lle} to visualize the gradients in these communication rounds in \cref{fig:cifar10_vis}.
From \cref{fig:cifar10_vis}, we observe that a group of densely distributed honest gradients (blue circles) skew away from the optimal gradient (green stars).
We call these blue circles \emph{"skewed gradients"} and name this phenomenon \emph{"gradient skew"}.
We provide visualization results on more datasets in \cref{appsec:vis_results}.
From visualization results on different datasets, we observe that the "gradient skew" phenomenon is prevalent across different datasets.

\section{Circumvent Robust AGRs under Gradient Skew}
Inspired by the above observation of the gradient skew phenomenon, we design a novel attack principle that can exploit this phenomenon to circumvent robust AGRs -- 
\emph{hide Byzantine gradients in the densely distributed skewed gradients}.

A body of recent works \cite{farhadkhani2022reasm, karimireddy2022bucketing,allouah2023nnm} have formulated Byzantine resilience for general robust AGRs.
These formulations commonly show that Byzantine defenses inherently trust densely distributed gradients, regarding them as honest.
We take the definition of $(f, \kappa)$-robustness in \cite{allouah2023nnm} as an example.

\begin{definition}
[$(f, \kappa)$-robustness]
\label{def:resilience_nnm}
Let $f<n/2$ and $\kappa\ge0$,
a robust AGR $\gA$ is called $(f,\kappa)$-robust] if 
for any input $\set{\vg_1,\ldots,\vg_n}$ and any set $\gG\subseteq\set{1,\ldots,n}$ of size $n-f$, the output $\hvg$ of AGR $\gA$ satisfies:
\begin{equation}
\begin{gathered}
\norm{\gA(\vg_1,\ldots,\vg_n)-\bvg_\gG}^2\le\frac{\kappa}{n-f}\sum_{i\in\gS} \norm{\vg_i-\bvg_\gG}^2,
\\\text{where}\quad
\bvg_\gG=\sum_{i\in\gG}\vg_i/(n-f).
\end{gathered}
\end{equation}
\end{definition}

In the definition, the distance between the aggregated gradients $\hvg$ and average candidate gradient $\bvg_\gG$ is upper-bounded by $\sum_{i\in\gS} \norm{\vg_i-\bvg_\gG}^2$, which measures the distribution density of gradients.
This means that the aggregated gradient is biased to densely distributed gradients.
In other words, \emph{Byzantine defenses believe that the most densely distributed $n-f$ gradients are more likely to be the honest ones}.

This inductive bias can be exploited by a malicious party when gradient skew exists.
In particular, we propose to \emph{hide Byzantine gradients in the skewed gradients}, i.e., place pink cross within blue dots as shown in \cref{fig:main_visualization}, to fake Byzantine gradients as honest ones.
As shown in \cref{fig:main_visualization}, Byzantine gradients and skew gradients, i.e., pink cross and blue dots, are densely distributed. 
As a result, they would be mistaken as honest gradients by Byzantine defenses.

This attack strategy enjoys another advantage.
It tricks Byzantine defense into thinking other outlier honest gradients, i.e., gray circles in \cref{fig:main_visualization}, are malicious.
As a result, these outlier gradients would be assigned lower weights or even removed from aggregation.
Unfortunately, outlier gradients are crucial to improving the generalization performance of the final FL model \cite{yan2024recess}.
As a result, this attack strategy can pose a significant threat to model performance.

\begin{figure*}
    \centering
    \begin{subfigure}[t]{0.45\textwidth}
        \centering
        \includegraphics[width=0.93\textwidth]{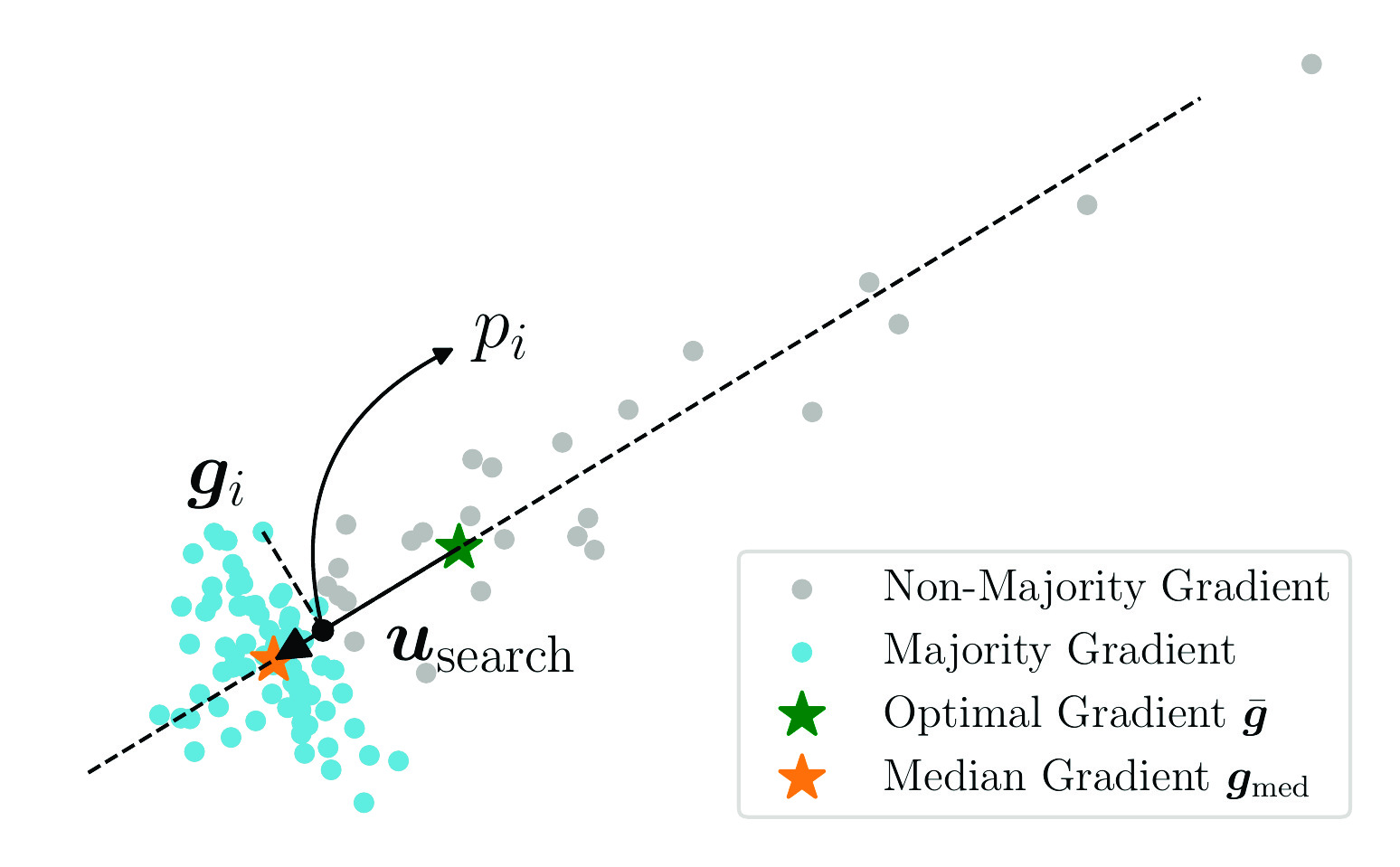}
        \caption{We search along the direction $\vu_\text{search}=\vg_\text{med}-\bvg$.
        The honest gradients with the largest scalar projection $p_i$ are selected as the skewed honest gradients (blue circles).}
        \label{fig:search}
    \end{subfigure}
    \hspace{0.05\linewidth}
    \begin{subfigure}[t]{0.45\textwidth}
        \centering
        \includegraphics[width=0.93\textwidth]{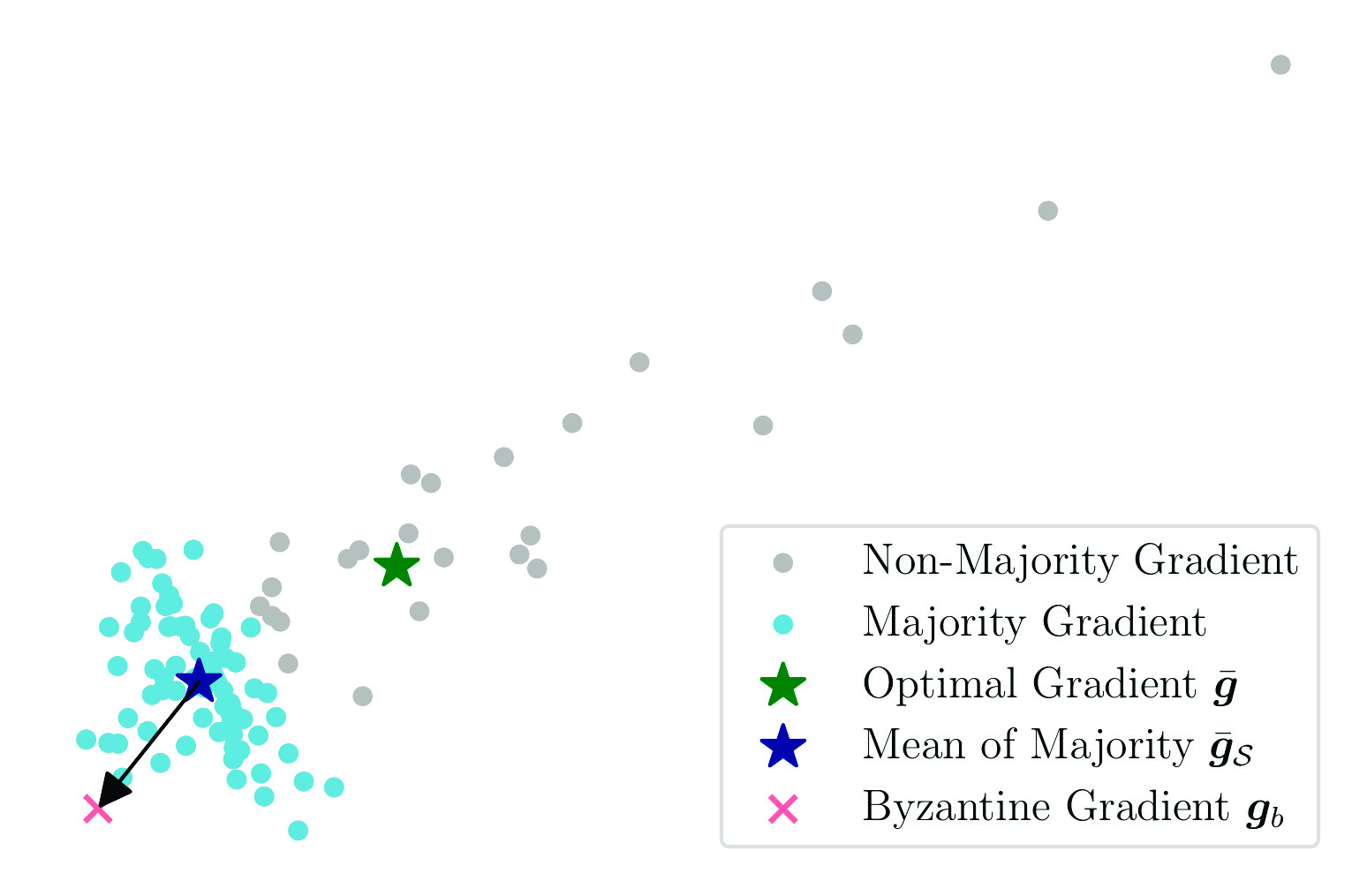}
        \caption{We start from the average of skewed honest gradients $\bvg_\gS$ (dark blue star) and select $\alpha$ such that Byzantine gradient $\vg_b$ (pink cross) lies within the skewed honest gradients.}
        \label{fig:hide}
    \end{subfigure}
    \caption{Illustration of the proposed two-stage attack \SKEW:
    in the first stage, \SKEW searches for the skewed honest gradients;
    in the second stage, \SKEW hides Byzantine gradients within the skewed honest gradients.
    }
    \label{fig:attack}
\end{figure*}

\section{Proposed Attack}
In this section, we introduce the proposed \abbreviation called \SKEW.
As discussed in the previous section, the attack principle is to \emph{hide Byzantine gradients within the skewed gradients}.
To achieve this goal, we carry out \SKEW attack in two stages:
in the first stage, we search for the skewed honest gradients;
in the second stage, we construct Byzantine gradients within the skewed honest gradients found in the first stage.
The procedure of \SKEW attack is shown in \cref{alg:proposed} in \cref{appsec:algorithm}.

\textbf{Search for the skewed honest gradients.}
To hide the Byzantine gradient in the skewed honest gradients, we first need to find the skewed honest gradients.
We perform a heuristic search motivated by Karl Pearson's formula \cite{knoke2002pearson2,moore2009pearson1}.
\cref{fig:search} illustrates the search procedure in this stage.

As visualized in \cref{fig:main_visualization}, skewed honest gradients are densely distributed.
As the population mode (in statistics) falls where the probability density is highest, the skewed honest gradients coincide with the (population) mode.
Thus, we can \emph{identify honest gradients near the mode as skewed gradients}.

Karl Pearson's formula \cite{knoke2002pearson2,moore2009pearson1} implies that the mode and median lie on the same side of the mean.
Therefore, the search for the skewed gradients starts from the mean and advances towards the median.
That is, as shown in \cref{fig:search} we search for the skewed honest gradients along the direction $\vu_\text{search}$ defined as:
\begin{align}
\vu_\text{search}=\vg_{\text{med}}-\bvg,
\end{align}
where $\vg_{\text{med}}$ is the coordinate-wise median of honest gradients $\set{\vg_i\mid i\in\gH}$, 
i.e., the $k$-th coordinate of $\vg_{\text{med}}$ is  $(\vg_\text{med})_k=\text{median}\{(\vg_i)_k\mid i\in\gH\}$, $\text{median}\{\cdot\}$ returns the median of the input numbers,
and $\bvg=\sum_{i\in\gH}\vg_i/(n-f)$ is the average of honest gradients.

For each honest gradient $\vg_i$, we compute its scalar projection $p_i$ on the searching direction $\vu_\text{search}$:
\begin{align}
p_i = \ip{\vg_i}{\frac{\vu_\text{search}}{\norm{\vu_\text{search}}}},
\quad\forall i\in\gH,
\end{align}
where $\ip{\cdot}{\cdot}$ represents the inner product.
The $n-2f$ gradients with the highest scalar projection values are identified as the skewed honest gradients.
The goal is to have AGR consider the selected $n-2f$ gradients as honest and the unselected $f$ gradients as Byzantine.
Let $\gS$ denote the index set, that is
\begin{equation}
\begin{gathered}
\gS=\text{Set of }(n-2f)\text{ indices of the gradients with the}\\
\text{highest scalar projection }p_i,
\end{gathered}
\end{equation}
then the skewed honest gradients are $\set{\vg_i\mid i\in\gS}$. 

\textbf{Hide Byzantine gradients within the skewed honest gradients.}
In this stage, we aim to hide Byzantine gradients $\set{\vg_i\mid i\in\gB}$ within the skewed honest gradients $\set{\vg_i\mid i\in\gS}$ identified in stage 1.
The primary goal of our attack is to disguise Byzantine gradients and the skewed honest gradients $\set{\vg_i\mid i\in\gB\cup\gS}$ as honest gradients.
Meanwhile, the secondary goal is to maximize the attack effect, i.e.,  maximize the distance between these "fake" honest gradients and the optimal gradient.
The hiding procedure in this stage is illustrated in \cref{fig:hide}.

According to \cite{farhadkhani2022reasm}, robust AGRs are sensitive to the diameter of gradients.
Therefore, we ensure that the Byzantine gradients $\vg_b$ lie within the diameter of the skewed honest gradients $\vg_s$ in order not to be detected.
\begin{align}
\label{eq:hide}
\norm{\vg_b-\vg_s}\le\max_{i,j\in\gS}{\norm{\vg_i-\vg_j}},\quad\forall b\in\gB,s\in\gS,
\end{align}
where $\gB$ is the index set of Byzantine clients, $\gS$ is the index set of skewed honest clients.

Meanwhile, we want to maximize the attack effect.
As Byzantine defenses assume densely distributed gradients, i.e., the skewed honest gradients $\set{\vg_s\mid s\in\gS}$ and Byzantine gradients $\set{\vg_b\mid s\in\gB}$, to be honest.
The aggregated gradients would be close to the mean of densely distributed gradients $\bvg_{\gS\cup\gB}=\sum_{i\in\gS\cup\gB}\vg_i/(n-f)$.
Therefore, we maximize the distance between $\bvg_{\gS\cup\gB}$ and the optimal gradient $\bvg$ to maximize the attack effect.
\begin{align}
\label{eq:constraint}
\max_{\set{\vg_b\mid b\in\gB}}{\norm{\bvg_{\gS\cup\gB}-\bvg}}.
\end{align}

Combining \cref{eq:hide} and \cref{eq:constraint}, our objective can be formulated as the following constrained optimization problem.
\begin{equation}
\begin{aligned}
\label{eq:optimization}
&\max_{\set{\vg_b\mid b\in\gB}}{\norm{\bvg_{\gS\cup\gB}-\bvg}}\\
\text{s.t.}\quad
&\bvg_{\gS\cup\gB}=\sum_{i\in\gS\cup\gB}\vg_i/(n-f)\\
&\norm{\vg_b-\vg_s}\le\max_{i,j\in\gS}{\norm{\vg_i-\vg_j}},\,\forall b\in\gB,s\in\gS
\end{aligned}
\end{equation}
\cref{eq:optimization} is too complex to be solved due to the high complexity of its feasible region.
Therefore, we restrict $\set{\vg_b\mid b\in\gB}$ to the following form:
\begin{align}
\label{eq:restrict}
\vg_b=\bvg_\gS+\alpha\cdot\sign(\bvg_\gS-\bvg)\odot\vsigma_\gS,
\quad\forall b\in\gB,
\end{align}
where
$\bvg_\gS=\sum_{i\in\gS}\vg_i/(n-2f)$ is the average of the skewed honest gradients,
$\alpha$ is a non-negative real number that controls the attack strength,
$\sign(\cdot)$ returns the element-wise indication of the sign of a number,
$\odot$ is the element-wise multiplication,
and $\vsigma_\gS$ is the element-wise standard deviation of skewed honest gradients $\set{\vg_i\mid i\in\gS}$.
$\bvg_\gS$ lies within the feasible region of \cref{eq:optimization}, which ensures that $\set{\vg_b\mid b\in\gB}$ are feasible when $\alpha=0$.
$\sign(\bvg_\gS-\bvg)$ controls the element-wise attack direction, and ensures that $\vg_b$ is farther away from the optimal gradient $\bvg$ under a larger $\alpha$.
$\vsigma_\gS$ controls the element-wise attack strength and ensures that Byzantine gradients are covert in each dimension.
\begin{table*}[t]
\begin{center}
{
\small

\begin{tabular}{lccccccc}																													
\toprule																													
\multicolumn{8}{c}{CIFAR-10}																													\\
\midrule																													
Attack	&	Multi-Krum			&	Median			&	RFA			&	Aksel			&	CClip			&	DnC			&	RBTM			\\
\midrule																													
BitFlip	&	54.76	$\pm$	0.06	&	53.73	$\pm$	2.05	&	56.04	$\pm$	3.13	&	51.99	$\pm$	2.04	&	54.44	$\pm$	0.46	&	60.81	$\pm$	0.56	&	55.21	$\pm$	3.72	\\
LIE	&	57.89	$\pm$	0.22	&	49.20	$\pm$	3.27	&	53.90	$\pm$	5.43	&	46.73	$\pm$	4.86	&	63.11	$\pm$	0.43	&	61.58	$\pm$	2.85	&	58.84	$\pm$	0.64	\\
IPM	&	47.55	$\pm$	1.75	&	51.68	$\pm$	1.85	&	55.36	$\pm$	2.10	&	56.85	$\pm$	2.07	&	58.75	$\pm$	5.59	&	62.30	$\pm$	3.60	&	48.43	$\pm$	0.17	\\
MinMax	&	59.44	$\pm$	3.41	&	57.27	$\pm$	0.63	&	60.20	$\pm$	1.63	&	57.17	$\pm$	5.50	&	59.38	$\pm$	5.15	&	62.53	$\pm$	2.67	&	57.72	$\pm$	2.94	\\
MinSum	&	55.47	$\pm$	1.70	&	52.27	$\pm$	0.53	&	54.59	$\pm$	2.38	&	56.43	$\pm$	1.74	&	54.70	$\pm$	1.96	&	61.89	$\pm$	1.62	&	46.78	$\pm$	0.32	\\
Mimic	&	56.00	$\pm$	4.26	&	52.55	$\pm$	0.89	&	53.61	$\pm$	0.86	&	57.19	$\pm$	2.50	&	51.00	$\pm$	0.11	&	62.10	$\pm$	5.22	&	46.77	$\pm$	2.52	\\
\SKEW (Ours)	&	\textbf{42.90}	$\pm$	1.97	&	\textbf{48.29}	$\pm$	0.40	&	\textbf{52.92}	$\pm$	1.75	&	\textbf{38.31}	$\pm$	0.47	&	\textbf{50.67}	$\pm$	0.27	&	\textbf{59.16}	$\pm$	1.84	&	\textbf{44.82}	$\pm$	0.97	\\
\midrule																													
\multicolumn{8}{c}{ImageNet-12}																													\\
\midrule																													
Attack	&	Multi-Krum			&	Median			&	RFA			&	Aksel			&	CClip			&	DnC			&	RBTM			\\
\midrule																													
BitFlip	&	59.62	$\pm$	0.73	&	58.56	$\pm$	4.80	&	59.71	$\pm$	5.00	&	61.64	$\pm$	1.98	&	14.87	$\pm$	1.58	&	59.78	$\pm$	1.50	&	58.49	$\pm$	1.99	\\
LIE	&	62.66	$\pm$	0.30	&	51.41	$\pm$	1.52	&	60.99	$\pm$	1.22	&	54.14	$\pm$	3.14	&	16.19	$\pm$	3.95	&	67.85	$\pm$	2.87	&	67.12	$\pm$	0.39	\\
IPM	&	52.66	$\pm$	2.01	&	59.20	$\pm$	2.44	&	61.25	$\pm$	0.62	&	59.17	$\pm$	1.27	&	14.33	$\pm$	5.95	&	66.31	$\pm$	3.60	&	55.93	$\pm$	0.57	\\
MinMax	&	68.17	$\pm$	1.91	&	67.76	$\pm$	0.07	&	63.05	$\pm$	0.75	&	59.33	$\pm$	3.85	&	20.99	$\pm$	3.07	&	68.05	$\pm$	1.59	&	65.99	$\pm$	1.26	\\
MinSum	&	57.50	$\pm$	3.09	&	58.78	$\pm$	2.10	&	64.04	$\pm$	0.69	&	67.15	$\pm$	0.32	&	16.38	$\pm$	2.70	&	68.69	$\pm$	1.18	&	61.70	$\pm$	1.62	\\
Mimic	&	66.86	$\pm$	0.04	&	59.39	$\pm$	6.07	&	60.45	$\pm$	7.09	&	58.94	$\pm$	1.27	&	11.35	$\pm$	2.26	&	69.07	$\pm$	4.69	&	55.26	$\pm$	1.30	\\
\SKEW (Ours)	&	\textbf{27.24}	$\pm$	1.63	&	\textbf{42.98}	$\pm$	1.62	&	\textbf{43.30}	$\pm$	3.13	&	\textbf{38.11}	$\pm$	1.02	&	\textbf{8.33}	$\pm$	1.85	&	\textbf{53.40}	$\pm$	4.94	&	\textbf{38.81}	$\pm$	0.65	\\
\midrule																													
\multicolumn{8}{c}{FEMNIST}																													\\
\midrule																													
Attack	&	Multi-Krum			&	Median			&	RFA			&	Aksel			&	CClip			&	DnC			&	RBTM			\\
\midrule																													
BitFlip	&	82.67	$\pm$	5.13	&	71.57	$\pm$	3.61	&	83.41	$\pm$	4.33	&	81.42	$\pm$	3.45	&	83.85	$\pm$	8.50	&	83.58	$\pm$	5.20	&	82.58	$\pm$	6.08	\\
LIE	&	68.11	$\pm$	6.86	&	58.38	$\pm$	7.06	&	66.19	$\pm$	7.93	&	38.48	$\pm$	3.32	&	73.03	$\pm$	3.86	&	77.42	$\pm$	5.60	&	53.35	$\pm$	5.17	\\
IPM	&	84.12	$\pm$	3.06	&	72.60	$\pm$	8.42	&	83.42	$\pm$	4.13	&	78.28	$\pm$	7.37	&	84.93	$\pm$	4.41	&	83.03	$\pm$	5.02	&	83.21	$\pm$	6.42	\\
MinMax	&	68.42	$\pm$	5.91	&	66.44	$\pm$	5.88	&	71.55	$\pm$	5.98	&	34.22	$\pm$	4.94	&	72.12	$\pm$	4.39	&	75.40	$\pm$	3.78	&	59.23	$\pm$	3.41	\\
MinSum	&	62.06	$\pm$	3.13	&	65.46	$\pm$	3.66	&	70.36	$\pm$	7.24	&	44.91	$\pm$	3.90	&	75.40	$\pm$	4.88	&	77.11	$\pm$	3.61	&	68.10	$\pm$	8.86	\\
Mimic	&	83.15	$\pm$	3.46	&	74.00	$\pm$	4.79	&	83.87	$\pm$	3.00	&	79.06	$\pm$	7.21	&	83.94	$\pm$	5.25	&	82.22	$\pm$	5.40	&	81.92	$\pm$	3.40	\\
\SKEW (Ours)	&	\textbf{22.13}	$\pm$	7.78	&	\textbf{55.19}	$\pm$	3.49	&	\textbf{39.43}	$\pm$	5.06	&	\textbf{16.58}	$\pm$	3.63	&	\textbf{18.88}	$\pm$	4.30	&	\textbf{17.56}	$\pm$	5.95	&	\textbf{39.33}	$\pm$	11.98	\\
\bottomrule																													
\end{tabular}					
}
\end{center}
\caption{
Accuracy (mean$\pm$std) under different attacks against different defenses on CIFAR-10, ImageNet-12, and FEMNIST.
The best attack performance is in bold (the \emph{lower}, the better).
}
\label{tbl:main_experiments}
\end{table*}

With the restriction in \cref{eq:restrict}, \cref{eq:optimization} can be simplified to the following optimization problem,
\begin{equation}
\label{eq:simplified}
\begin{gathered}
\max{\alpha}\\
\text{s.t.}\quad
\norm{\bvg_\gS+\alpha\cdot\sign(\bvg_\gS)\odot\vsigma_\gS-\vg_s}\le\max_{i,j\in\gS}{\norm{\vg_i-\vg_j}},\\
\forall s\in\gS,
\end{gathered}
\end{equation}
which can be easily solved by the bisection method described in \cref{appsec:bisection}.
While $\alpha$ that solves \cref{eq:simplified} is provable in most cases, we find in practice that a slightly adjusted attack strength can further improve the effect of \SKEW.
We use an additional hyperparameter $\nu(>0)$ to control the attack strength of \SKEW.
\SKEW sets $\vg_b=\bvg_\gS+\nu\alpha\cdot\sign(\bvg_\gS)\odot\vsigma_\gS-\vg_i$ for all $b\in\gB$ and uploads Byzantine gradients to the server.
Higher $\nu$ implies higher attack strength.
We discuss the performance of \SKEW with different $\nu$ in \cref{appsec:nus}.

\section{Experiments}
We conduct all experiments on the same workstation with 8 Intel(R) Xeon(R) Platinum 8336C CPUs, a NVIDIA Tesla V100, and 64GB main memory running Linux platform.

\subsection{Experimental Setups}
\label{subsec:experimental_setups}

\textbf{Datasets.} Our experiments are conducted on three real-world datasets: CIFAR-10 \cite{krizhevsky2009cifar}, a subset of ImageNet \cite{russakovsky2015imagenet} refered as ImageNet-12 \cite{li2021imagenet12} and FEMNIST \cite{caldas2018leaf}. Please refer to \cref{appsec:data_distribution} for more details about the data distribution.

More detailed setups are deferred to \cref{appsec:main_setups}.

\textbf{Baseline attacks.} 
We consider six state-of-the-art attacks:
BitFlip \cite{allen2020safeguard},
LIE \cite{baruch2019lie},
IPM \cite{xie2020ipm},
Min-Max \cite{shejwalkar2021dnc},
Min-Sum \cite{shejwalkar2021dnc},
and Mimic \cite{karimireddy2022bucketing}.
Among the above six attacks, BitFlip and LabelFlip are popular agnostic attacks;
LIE, Min-Max and Min-Sum are partial knowledge attacks;
IPM is an omniscient attack.
The detailed introduction and hyperparameter settings of these attacks are shown in \cref{appsec:attack_hyperparam}.

\begin{figure*}[t]
    \includegraphics[width=.24\linewidth,align=c]{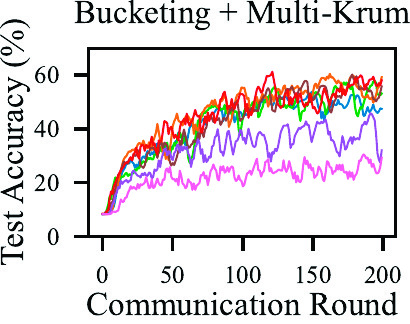}
    \includegraphics[width=.24\linewidth,align=c]{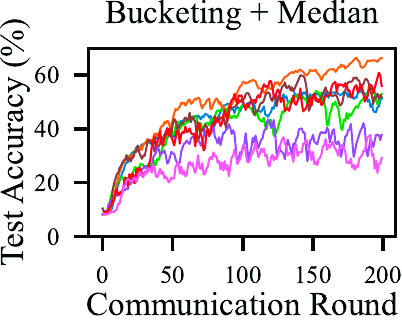}
    \includegraphics[width=.24\linewidth,align=c]{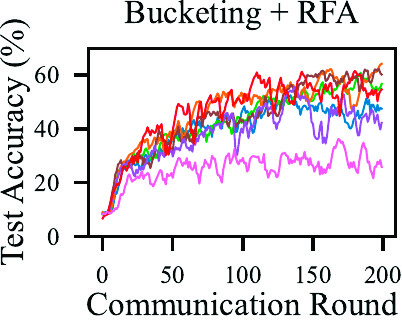}
    \includegraphics[width=.24\linewidth,align=c]{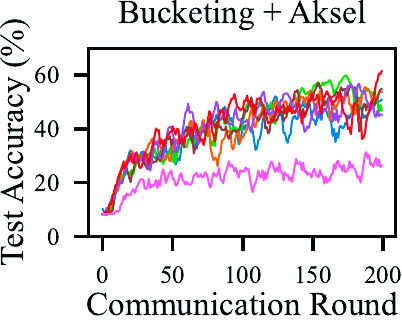}
    \\
    \hfill\\
    \includegraphics[width=.24\linewidth,align=c]{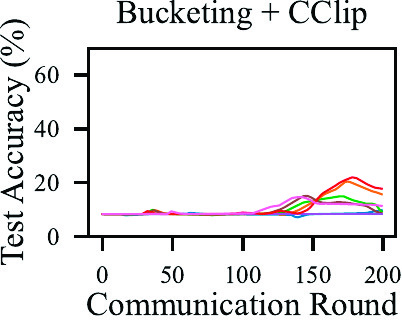}
    \includegraphics[width=.24\linewidth,align=c]{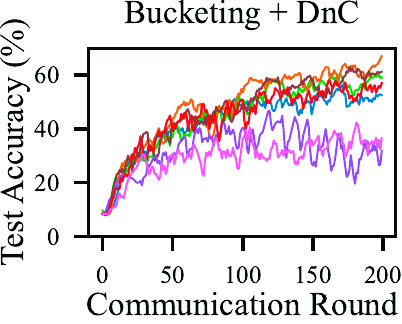}
    \includegraphics[width=.24\linewidth,align=c]{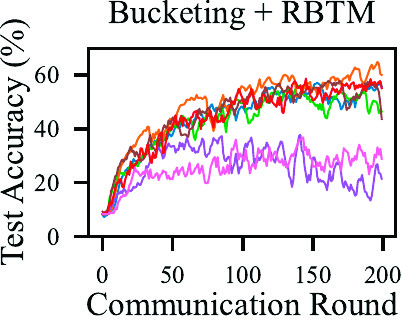}
    \hspace{.05\linewidth}
    \includegraphics[width=.15\linewidth,align=c]{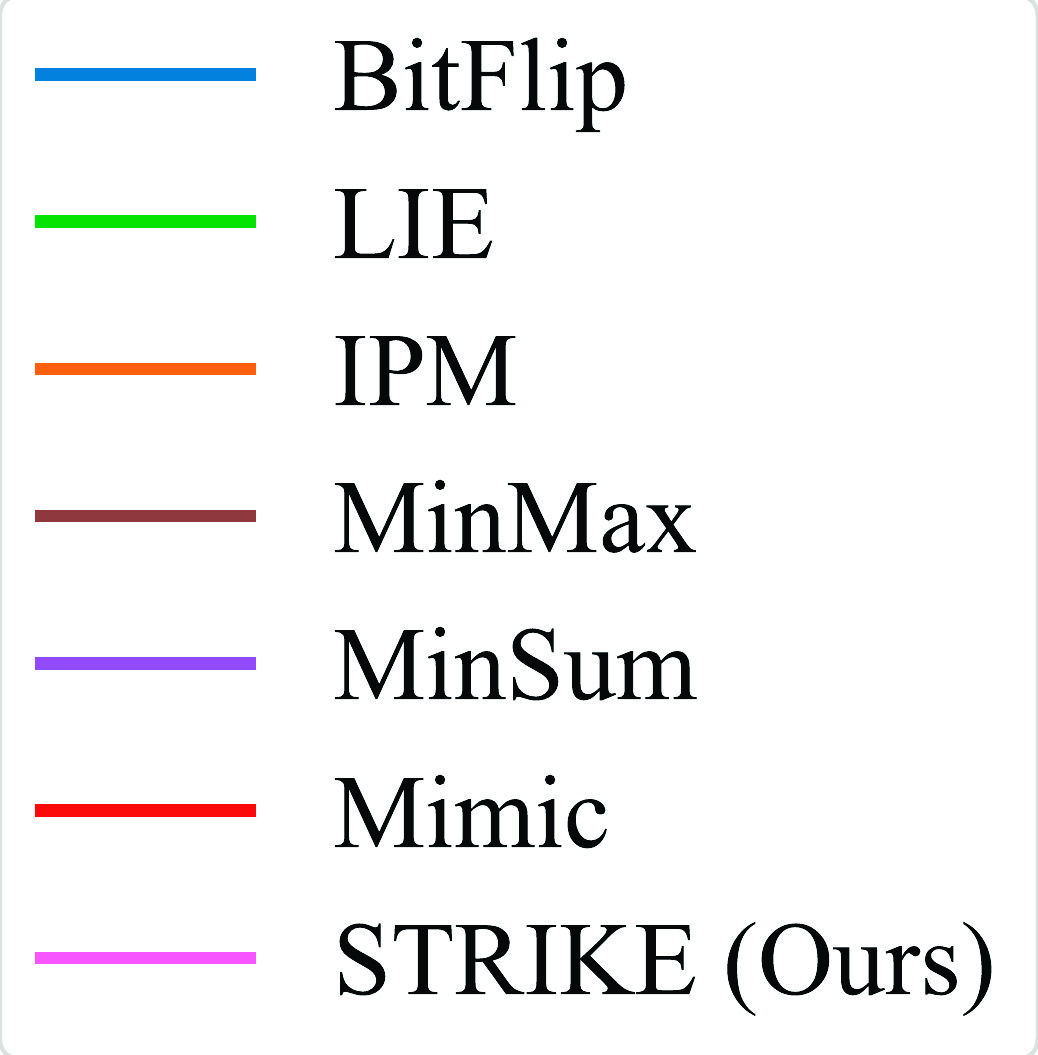}
    \\
    \caption{
    Accuracy under different attacks against seven robust AGRs with bucketing on ImageNet-12.
    The \emph{lower}, the better.
    }
    \label{fig:bucket_results}
\end{figure*}

\textbf{Evaluated defenses.}
We evaluate the performance of our attack on the following robust AGRs:
Multi-Krum \cite{blanchard2017krum},
Median \cite{yin2018mediantrmean},
RFA \cite{pillutla2019geometric},
Aksel \cite{boussetta2021aksel},
CClip \cite{karimireddy2021cc}
DnC \cite{shejwalkar2021dnc},
and RBTM \cite{el2021collaborative}.
Besides, we also consider bucketing \cite{karimireddy2022bucketing} and NNM \cite{allouah2023nnm}, two simple yet effective schemes that adapt existing robust AGRs to the non-IID setting.
The detailed hyperparameter settings of the above robust AGRs are listed in \cref{appsec:defense_hyperparams}.

\subsection{Experiment Results}
\label{subsec:results}

\textbf{Attacking against various robust AGRs.}
\cref{tbl:main_experiments} demonstrates the performance of seven different attacks against seven robust AGRs on CIFAR-10, ImageNet-12, and FEMNIST datasets.
From \cref{tbl:main_experiments}, we can observe that our \SKEW attack generally outperforms all the baseline attacks against various defenses on all datasets, verifying the efficacy of our \SKEW attack.
On ImageNet-12 and FEMNIST, the improvement of \SKEW over the best baselines is more significant.
We hypothesize that this is because the skew degree is higher on ImageNet-12 and FEMNIST compared to CIFAR-10.
Since \SKEW exploits gradient skew to launch Byzantine attacks, it is more effective on ImageNet-12 and FEMNIST.
DnC demonstrates almost the strongest resilience to previous baseline attacks.
This is because these attacks fail to be aware of the skew nature of honest gradients in FL.
By contrast, our \SKEW attack can take advantage of gradient skew and circumvent DnC defense.
The above observations clearly validate the superiority of \SKEW.

\textbf{Attacking against robust AGRs with bucketing.}
\cref{fig:bucket_results} demonstrates the performance of seven different attacks against the bucketing scheme \cite{karimireddy2022bucketing} with different robust AGRs.
The results demonstrate that our \SKEW attack works best against Multi-Krum, RFA, and Aksel.
When attacking against DnC, Median, and RBTM, only MinSum attack is comparable to our \SKEW attack.

\textbf{Attacking against robust AGRs with NNM.}
\cref{tab:nnm} compare the performance of STRKE attack against top-3 strongest attacks against the NNM scheme \cite{karimireddy2022bucketing} under the top-3 most robust robust AGRs.
The results suggest that the proposed STRIKE attack still outperforms other baseline attacks against NNM.

\begin{table}[t]
\centering
{\small
\begin{tabular}{lccc}							
\toprule							
Attack	&	NNM + Median	&	NNM + RFA	&	NNM + DnC	\\
\midrule							
BitFlip	&	57.14	&	58.55	&	53.68	\\
LIE	&	58.04	&	58.68	&	58.87	\\
Mimic	&	66.15	&	67.43	&	69.35	\\
STRIKE	&	\textbf{39.61}	&	\textbf{40.38}	&	\textbf{38.91}	\\
\bottomrule							
\end{tabular}							
}
\caption{Accuracy under different attacks against NNM on ImageNet-12. The best results are in bold (The \emph{lower}, the better).}
\label{tab:nnm}
\end{table}

\textbf{Imparct of $\nu$ on \SKEW attack.}
We study the influence of $\nu$ on ImageNet-12 dataset. We report the test accuracy under \SKEW attack with $\nu$ in $\{0.25 * i\mid i=1,\ldots,8\}$ against seven different defenses on ImageNet-12.
As shown in \cref{fig:nus_results}, the performance of \SKEW is generally competitive with varying $\nu$.
In most cases, simply setting $\nu=1$ can beat almost all the attacks (except for CClip, yet we observe that the performance is low enough to make the model useless).

\textbf{The effectiveness of \SKEW attack under different non-IID levels.}
We vary Dirichlet concentration parameter $\beta$ in $\set{0.1, 0.2, 0.5, 0.7, 0.9}$ to study how our attack behaves under different non-IID levels.
We additionally test the performance in the IID setting.
As shown in \cref{fig:niid_level}, the accuracy generally increases as $\beta$ decreases for all attacks.
The accuracy under our \SKEW attack is consistently lower than that of all the baseline attacks.
Besides, we also note that the accuracy gap between our \SKEW attack and other baseline attacks gets smaller when the non-IID level decreases.
We hypothesize the reason is that gradient skew becomes milder as the non-IID level decreases.
Even in the IID setting, our \SKEW attack is competitive compared to other baselines.

\textbf{The performance of \SKEW attack with different Byzantine client ratios.}
We vary the number of Byzantine clients $f$ in $\set{5, 10, 15, 20}$ and fix the total number of clients $n$ to be $50$.
In this way, Byzantine client ratio $f/n$ varies in $\set{0.1, 0.2, 0.3, 0.4}$ to study how our attack behaves under different Byzantine client ratios.
As shown in \cref{fig:byz_ratio}, the accuracy generally decreases as $f/n$ increases for all attacks.
The accuracy under our \SKEW attack is consistently lower than that under all the baseline attacks.

\section{Conclusion}
In this paper, we observe the existence of the gradient skew phenomenon due to non-IID data distribution in FL.
Based on the observation, we propose a novel attack called \SKEW that can exploit the vulnerability.
Generally, \SKEW hides Byzantine gradients within the skewed honest gradients.
To this end, \SKEW first searches for the skewed honest gradients, and then constructs Byzantine gradients within the skewed honest gradients by solving a constrained optimization problem.
Empirical studies on three real-world datasets confirm the efficacy of our \SKEW attack.
The STRIKE relies on the gradient skew phenomenon, which is closely related to non-IIDness of data distribution.
When the data is IID, the performance could be limited.
Therefore, defenses that can alleviate non-IID can potentially mitigate our STRIKE attack.
In our future works, we will explore potential defenses against this threat.

\section*{Ethical Statement}

The proposed skew-aware Byzantine attack \SKEW can present a threat to federated learning.
Our goal with this work is thus to preempt these harms and encourage Byzantine defenses that are robust to skew-aware attacks in the future.

\section*{Acknowledgements}
This work is funded by National Key Research and Development Project (Grant No: 2022YFB2703100) and by the Pioneer R\&D Program of Zhejiang (No.2024C01021).
This work is also sponsored by Sony AI.

\bibliography{aaai25}

\appendix
\onecolumn
\section{Visualization of Gradient Skew}
\label{appsec:vis}

In order to gain insight into the gradient distribution, we use Locally Linear Embedding (LLE) \footnote{
Compared to LLE, t-SNE \cite{van2008tsne} is a more popular visualization technique.
Since t-SNE adjusts Gaussian bandwidth to locally normalize the density of data points, t-SNE can not capture the distance information of data. 
However, gradient skew relies heavily on distance information.
Therefore, t-SNE is not appropriate for the visualization of gradient skew.
In contrast, LLE can preserve the distance information of data distribution.
} \cite{roweis2000lle} to visualize the gradients.
From the visualization results, we observe that the distribution of gradient is skewed throughout FL training process when the data across different clients is non-IID.
In this section, we first provide the detailed experimental setups of the observation experiments and then present the visualization results.

\subsection{Experimental Setups}
\label{appsec:vis_setting}

For CIFAR-10, we set the number of clients $n=100$ and the Dirichlet concentration parameter $\beta=0.1$.
For ImageNet-12, we set the number of clients $n=50$ and the Dirichlet concentration parameter $\beta=0.1$.
For FEMNIST, we adopt its natural data partition as introduced in \cref{subsec:experimental_setups}.
For all three datasets, we set the number of Byzantine clients $f=0$.
For CIFAR-10 and FEMNIST, we sample 100 clients to participate in training in each communication round.
More visualized gradients would help us capture the characteristic of gradient distribution.
For ImageNet-12, we sample 50 clients in each communication round.
This is because we train ResNet-18 on ImageNet-12 and LLE on 100 gradients of ResNet-18 would be intractable due to the high dimensionality.
Other setups align with \cref{apptable:default_setup}.

For LLE, we set the number of neighbors to be $k=0.1m$, where $m$ is the number of sampled clients, to capture both local and global geometry of gradient distribution.

\subsection{Gradient Visualization Results}
\label{appsec:vis_results}

On each dataset, we run FedAvg for $T$ communication round.
Among the total $T$ communication rounds, we randomly sample 6 rounds for visualization.
For each round, we use LLE to visualize all the gradients and the optimal gradient (the average of all gradients) in this round.
Please note that LLE is not linear.
Therefore, the optimal gradient after the LLE may not be the average of all uploaded gradients after LLE.
The visualization results are posted in \cref{fig:vis_results} below.
In \cref{fig:vis_results}, the substantial gradients skew away from the optimal gradient.
These results imply that the gradient distribution is skewed during the entire training process.

\begin{figure}[H]
    \centering
    \includegraphics[width=.8\linewidth]{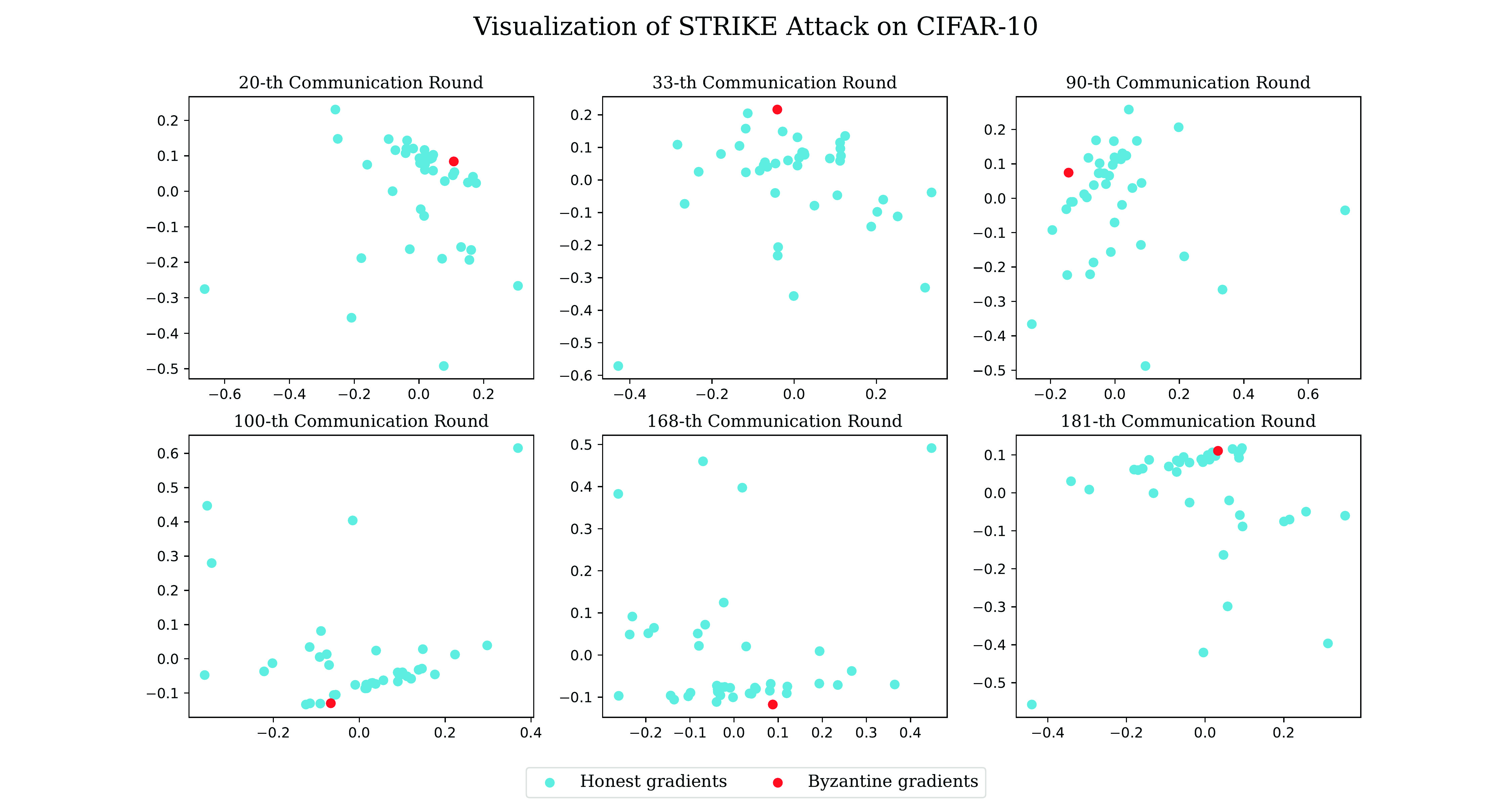}
    \caption{Visualization of STRIKE attack on CIFAR-10 datasets. The visualization shows that Byzantine gradients can hide within the skewed honest gradients well, which justifies that the heuristic search in the first stage of STRIKE attack can effectively find the skewed honest gradients.}
    \label{fig:pearson_skew}
\end{figure}

We also visualize the Byzantine gradients together with honest gradients under STRIKE attack against Median AGR on CIFAR-10 in the non-IID setting in \cref{fig:pearson_skew}.
The visualization shows that Byzantine gradients can hide within the skewed honest gradients well.
This justifies that the heuristic search in the first stage of STRIKE attack can effectively find the skewed honest gradients.
\begin{figure}[H]
    \centering
    \includegraphics[width=.9\linewidth]{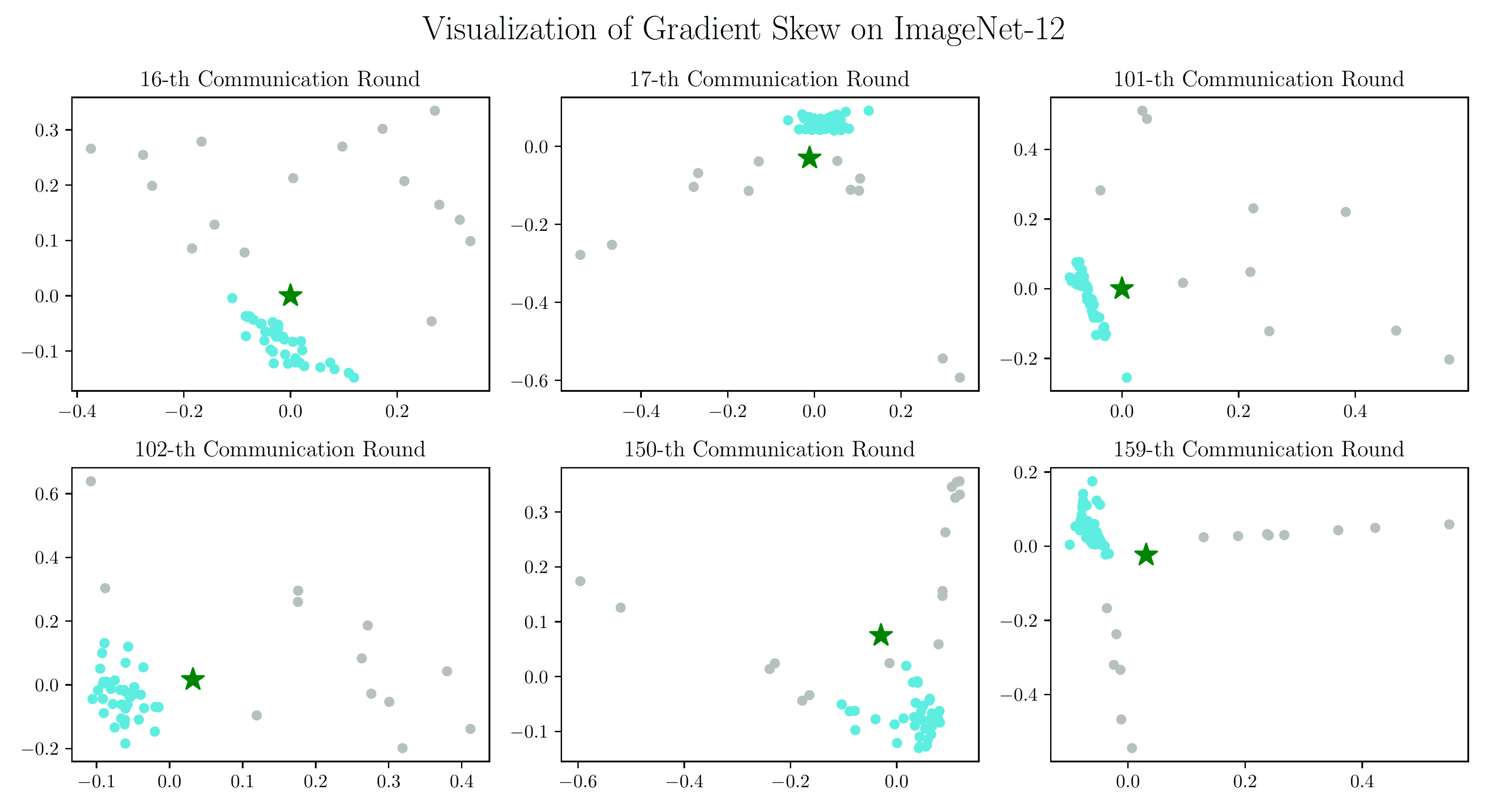}
    \includegraphics[width=.9\linewidth]{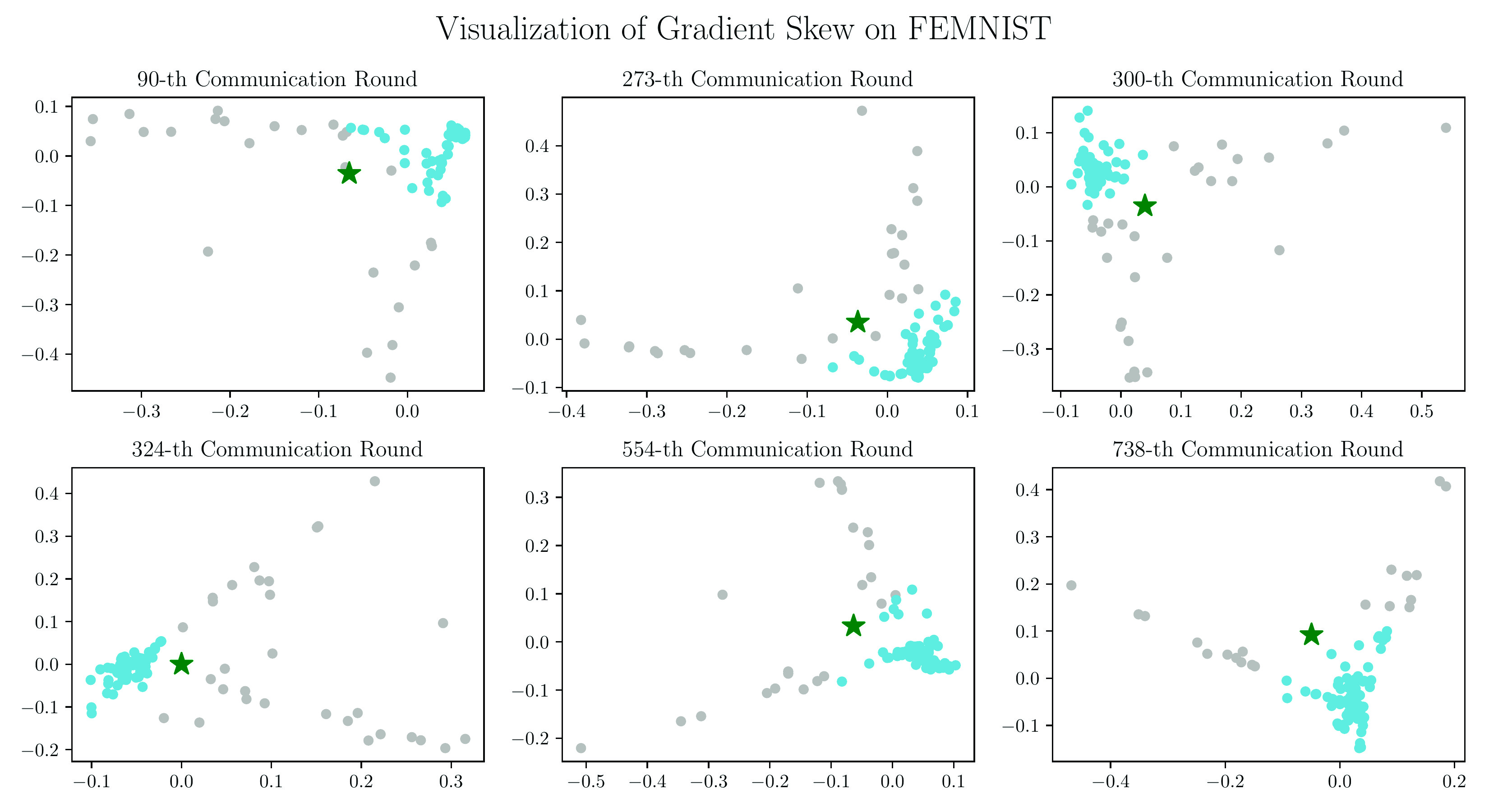}
    \includegraphics[width=.7\linewidth]{figures_cymk/new_vis_legend-1.jpg}
    \caption{Visualization of gradient skew on ImageNet-12 and FEMNIST}
    \label{fig:vis_results}
\end{figure}

\section{Algorithm of the Proposed STRKE Attack}
\label{appsec:algorithm}

\begin{algorithm}[H]
\caption{\SKEW Attack}
\label{alg:proposed}
\begin{algorithmic}
\REQUIRE Honest gradients $\set{\vg_i\mid i\in\gH}$,
hyperparameter $\nu>0$ that controls attack strength (default $\nu=1$)
\STATE $\vg_{\text{med}}\gets\text{Coordinate-wise median of }\set{\vg_i\mid i\in\gH}$
\COMMENT{\# Stage 1: search for the skewed densely distributed gradients}
\STATE $\vu_\text{search}\gets\vg_{\text{med}}-\bvg$
\FOR{$i\in\gH$}
    \STATE $p_i \gets \ip{\vg_i}{\vu_\text{search}/\norm{\vu_\text{search}}}$
\ENDFOR
\STATE $\gS\gets\text{Set of }n-f\text{ indices of honest gradients with the highest }p_i$
\STATE $\bvg_\gS\gets\sum_{i\in\gS}\vg_i/(n-2f)$
\COMMENT{\# Stage 2: hide Byzantine gradients within the skewed densely distributed gradients}
\STATE $\vsigma_\gS\gets\text{Coordinate-wise standard deviation of }\set{\vg_i\mid i\in\gS}$
\STATE solve \cref{eq:simplified} for $\alpha$
\FOR{$b\in\gB$}
    \STATE $\vg_b\gets\bvg_\gS+\nu\alpha\cdot\sign(\bvg_\gS-\bvg)\odot\vsigma_\gS$
\ENDFOR
\STATE \textbf{return} Byzantine gradients $\set{\vg_b\mid g\in\gB}$
\end{algorithmic}
\end{algorithm}

\section{Bisection Method to Solve \cref{eq:simplified}}
\label{appsec:bisection}
In this section, we present the bisection method used to solve \cref{eq:simplified}.
We define $f(\cdot)$ as follows.
\begin{align}
f(\alpha)=\max_{i\in\gS}\norm{\bvg_\gS+\alpha\cdot\sign(\bvg_\gS)\odot\vsigma_\gS-\vg_i}-\max_{i,j\in\gS}{\norm{\vg_i-\vg_j}},
\quad\alpha\in[0,+\infty).
\end{align}
We can easily verify the following facts: 1. $f(0)\le0$, $f(\alpha)\rightarrow+\infty$ when $\alpha\rightarrow+\infty$;
2. $f(\cdot)$ is continuous;
3. $f(\cdot)$ has unique zero point in $[0,+\infty)$.
Therefore, optimizing \cref{eq:simplified} is equivalent to finding the zero point of $f(\cdot)$, which can be easily solved by bisection method in \cref{alg:bisection}.

\begin{algorithm}[ht]
\caption{Bisection method}
\label{alg:bisection}
\begin{algorithmic}
\REQUIRE The skewed densely distributed honest gradients $\set{\vg_i\mid i\in\gS}$, tolerance $\varepsilon>0$, max iteration $M>0$
\STATE $\alpha_{\min}\gets0$
\STATE $\alpha_{\max}\gets1$
\WHILE{$f(\alpha)<0$}
\STATE $\alpha_{\max}\gets2\alpha_{\max}$
\ENDWHILE
\STATE $iter\gets0$
\WHILE{$\alpha_{\max}-\alpha_{\min}>\varepsilon$ and $iter<M$}
\STATE $\alpha_\text{mid}\gets(\alpha_{\max}+\alpha_{\min})/2$
\IF{$f(\alpha_\text{mid})<0$}
\STATE $\alpha_{\min}\gets\alpha_\text{mid}$
\ELSE
\STATE $\alpha_{\max}\gets\alpha_\text{mid}$
\ENDIF
\STATE $iter\gets iter+1$
\ENDWHILE
\STATE $\alpha\gets(\alpha_{\max}+\alpha_{\min})/2$
\STATE \textbf{return} $\alpha$
\end{algorithmic}
\end{algorithm}

Theoretically, the final optimization problem of the proposed STRIKE attack in Eq. (19) can be effectively solved by a bisection algorithm as discussed in Appendix C. The computation cost is $\mathcal{O}(-\log\epsilon)$, where $\epsilon$ is the error of $\alpha$. In experiments, we perform bisection only 8 times and make the error of $\alpha$ within $1\%$. Empirically, the attack time for STRIKE is 12.11s (13.47s for MinMax, 13.35s for MinSum) on ImageNet-12.

In contrast, benign clients perform local updates to compute local gradients. The computation cost depends on the local data size, model architecture, batch size, number of local epochs, etc. In our setting, the average local update time is 15.14s on CIFAR-10, 11.76s on ImageNet-12 and 27.13s on FEMNIST. Tests are performed on a single A100 GPU.

\section{Experimental Setups and Additional Experiments}
\label{appsec:experiments}
\subsection{Experimental Setups}
\label{appsec:main_setups}
\subsubsection{Data Distribution}
\label{appsec:data_distribution}
For CIFAR-10 \cite{krizhevsky2009cifar} and ImageNet-12, we use Dirichlet distribution to generate non-IID data by following \cite{yurochkin2019bayesian, li2021federated}. 
For each class $c$, we sample $\vq_c\sim\text{Dir}_n(\beta)$ and allocate a $(\vq_c)_i$ portion of training samples of class $c$ to client $i$.
Here, $\text{Dir}_n(\cdot)$ denotes the $n$-dimensional Dirichlet distribution, and $\beta>0$ is a concentration parameter.
We follow \cite{li2021federated} and set the number of clients $n=50$ and the concentration parameter $\beta=0.5$ as default.

For FEMNIST, the data is naturally partitioned into 3,597 clients based on the writer of the digit/character.
Thus, the data distribution across different clients is naturally non-IID.
For each client, we randomly sample a 0.9 portion of data as the training data and let the remaining 0.1 portion of data be the test data following \cite{caldas2018leaf}.

\subsubsection{Hyperparameter Setting of Baselines Attacks}
\label{appsec:attack_hyperparam}
The compared baseline attacks are: 
BitFlip \cite{allen2020safeguard},
LIE \cite{baruch2019lie},
IPM \cite{xie2020ipm},
Min-Max \cite{shejwalkar2021dnc},
Min-Sum \cite{shejwalkar2021dnc},
and Mimic \cite{karimireddy2022bucketing}.
\begin{itemize}
    \item \textbf{BitFlip.} A Byzantine client sends $-\nabla\gL_i(\vw_t)$ instead of $\nabla\gL_i(\vw_t)$ to the server.
    \item \textbf{LabelFlip.} Corrupt the datasets by transforming labels by $\mathcal{T}(y)=c-1-y$, where $c$ is the number of classes, $y$ is the label of a sample.
    \item \textbf{LIE.} The Byzantine clients estimate the mean $\vmu$ and element-wise standard deviation $\vsigma$ of the honest gradients, and send $\vmu-z\vsigma$ to the server, where $z$ controls the attack strength.
    \item \textbf{Min-Max.} The Byzantines first estimate the mean $\vmu$ of the honest gradients and select a perturbation vector $\vdelta$.
    Then, the Byzantines find scaling coefficient $\gamma$ by solving a maximizing-maximum-distance optimization problem and send $\vmu+\gamma\vdelta$ to the server.
    \item \textbf{Min-Sum.} Similar to Min-Max, but the Byzantine clients solve a maximizing-distance-sum optimization problem to determine the scaling coefficient $\gamma$ instead.
    \item \textbf{IPM.} The Byzantines send $-\varepsilon\bvg^t$ to the server, where $\varepsilon$ controls the attack strength.
    The Byzantines evaluate $\varepsilon$ based on the honest gradients and the server-side robust AGR to maximize the attack effect.
\end{itemize}
The hyperparameter setting of the above attacks is listed in the following table.
\begin{table}[ht]
\caption{
The hyperparameter setting of six baseline attacks. 
N/A represents there is no hyperparameter required for this attack.
}
\label{apptbl:attack_hyperparams}
\begin{center}
\begin{tabular}{ll}
\toprule
Attacks & Hyperparameters \\
\midrule
BitFlip & N/A \\
LIE & $z=1.5$ \\
IPM & $\varepsilon=0.1$\\
Min-Max & $\gamma_{\text{init}}=10,\tau=1\times10^{-5}$, $\nabla^p$: coordinate-wise standard deviation\\
Min-Sum & $\gamma_{\text{init}}=10,\tau=1\times10^{-5}$, $\nabla^p$: coordinate-wise standard deviation \\
Mimic & N/A \\
\bottomrule
\end{tabular}
\end{center}
\end{table}

\subsubsection{The Hyperparameter Setting of Evaluated Defenses}
\label{appsec:defense_hyperparams}
The performance of our attack is evaluated on seven recent robust defenses:
Multi-Krum \cite{blanchard2017krum},
Median \cite{yin2018mediantrmean},
RFA \cite{pillutla2019geometric},
Aksel \cite{boussetta2021aksel},
CClip\cite{karimireddy2021cc}
DnC \cite{shejwalkar2021dnc},
and RBTM \cite{el2021collaborative}.
The hyperparameter setting of the above defenses is listed in the following table.
\begin{table}[ht]
\caption{
The hyperparameter setting of seven evaluated defenses.
N/A represents there is no hyperparameter required for this defense.
}
\label{apptbl:defense_hyperparams}
\begin{center}
\begin{tabular}{ll}
\toprule
Defenses & Hyperparameters \\
\midrule
Multi-Krum & N/A \\
Median & N/A \\
RFA & $T=8$ \\
Aksel & N/A \\
CClip & $L=1,\tau=10$ \\
DnC & $c=1, \textsf{niters}=1, b=1000$ \\
RBTM & N/A \\
\bottomrule
\end{tabular}
\end{center}
\end{table}
we also consider a simple yet effective bucketing scheme \cite{karimireddy2022bucketing} that adapts existing defenses to the non-IID setting.
We follow the original paper and set the bucket size to be $s=2$.

\subsubsection{Evaluation}
We use top-1 accuracy, i.e., the proportion of correctly predicted testing samples to total testing samples, to evaluate the performance of global models.
The \emph{lower} the accuracy, the more effective the attack.
We run each experiment five times and report the mean and standard deviation of the highest accuracy during the training process.

\subsubsection{Compute}
All experiments are run on the same machine with Intel E5-2665 CPU, 32GB RAM, and four GeForce GTX 1080Ti GPU.

\subsubsection{Other Setups}
\label{appsec:other_setting}
The number of Byzantine clients of all datasets is set to $f=0.2\cdot n$.
We test \SKEW with $\nu\in\set{ 0.25\cdot i\mid i=1,\ldots,8}$ and report the lowest test accuracy (highest attack effectiveness).

The hyperparameter setting for datasets FEMNIST \cite{caldas2018leaf}, CIFAR-10 \cite{krizhevsky2009cifar} and ImageNet-12 \cite{russakovsky2015imagenet} are listed in below \cref{apptable:default_setup}.
\begin{table}[H]
\caption{Hyperparameter setting for FEMNIST, CIFAR-10 and ImageNet-12.
\# is the number sign. For example, \# Communication rounds represents the number of communication rounds.}
\label{apptable:default_setup}
\begin{center}
\begin{tabular}{llll}
\toprule 
Dataset & FEMNIST & CIFAR-10 & ImageNet-12 \\
Architecture & \makecell[l]{CNN \\ \citeauthor{caldas2018leaf}} & \makecell[l]{AlexNet \\ \cite{krizhevsky2017alexnet}} & \makecell[l]{ResNet-18 \\ \cite{he2016resnet}} \\
\midrule
\makecell[l]{\# Communication \\ rounds} & 800 & 200 & 200 \\
\# Sampled Clients & 10 & 50 & 50 \\
\midrule
\# Local epochs & 1 & 1 & 1 \\
Optimizer & SGD & SGD & SGD \\
Batch size & 128 & 128 & 128 \\
Learning rate & 0.5 & 0.1 & 0.1 \\
Momentum & 0.5 & 0.9 & 0.9 \\
Weight decay & 0.0001 & 0.0001 & 0.0001 \\
Gradient clipping & Yes & Yes & Yes \\
Clipping norm & 2 & 2 & 2 \\
\bottomrule
\end{tabular}
\end{center}
\end{table}


\subsection{Additional Experiments}

\subsubsection{Performance under Varying Hyperparameter $\nu$}
\label{appsec:nus}
\begin{figure}[ht]
    \includegraphics[width=.24\linewidth, align=c]{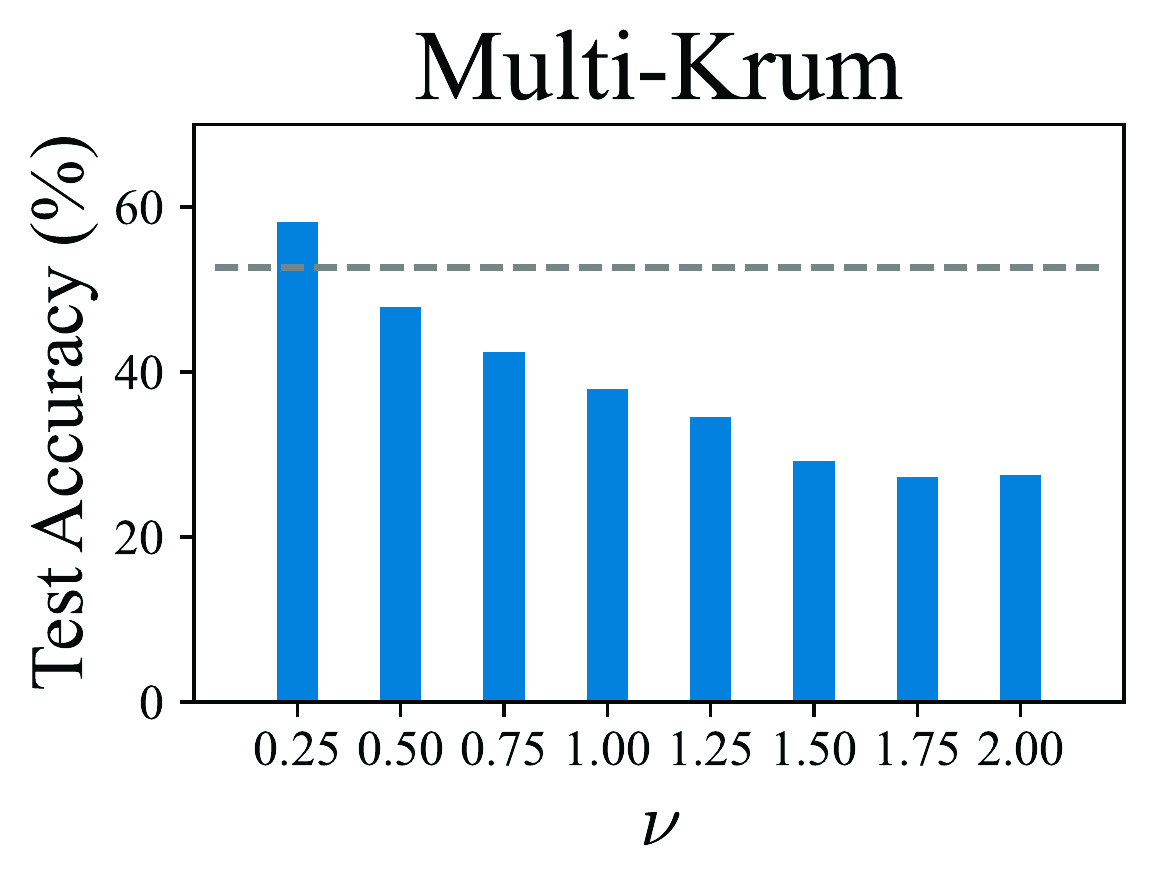}
    \includegraphics[width=.24\linewidth, align=c]{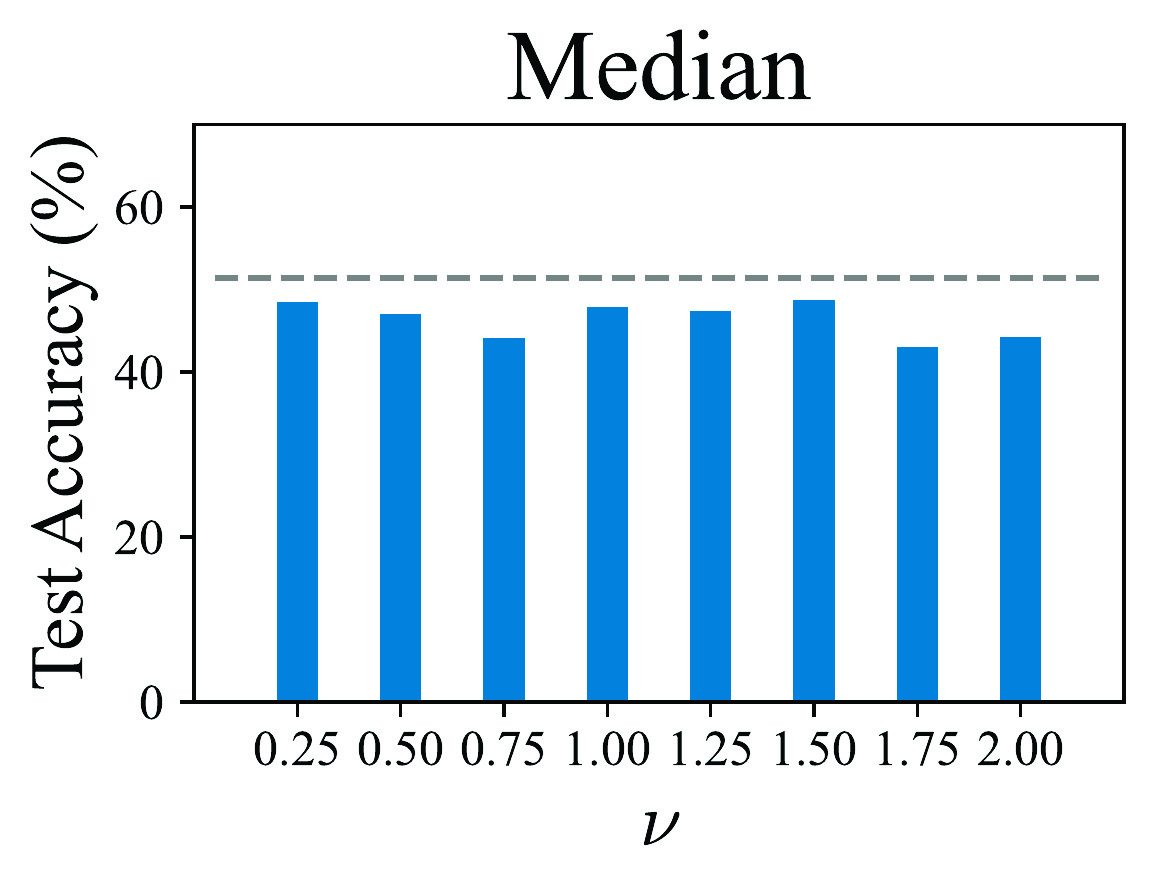}
    \includegraphics[width=.24\linewidth, align=c]{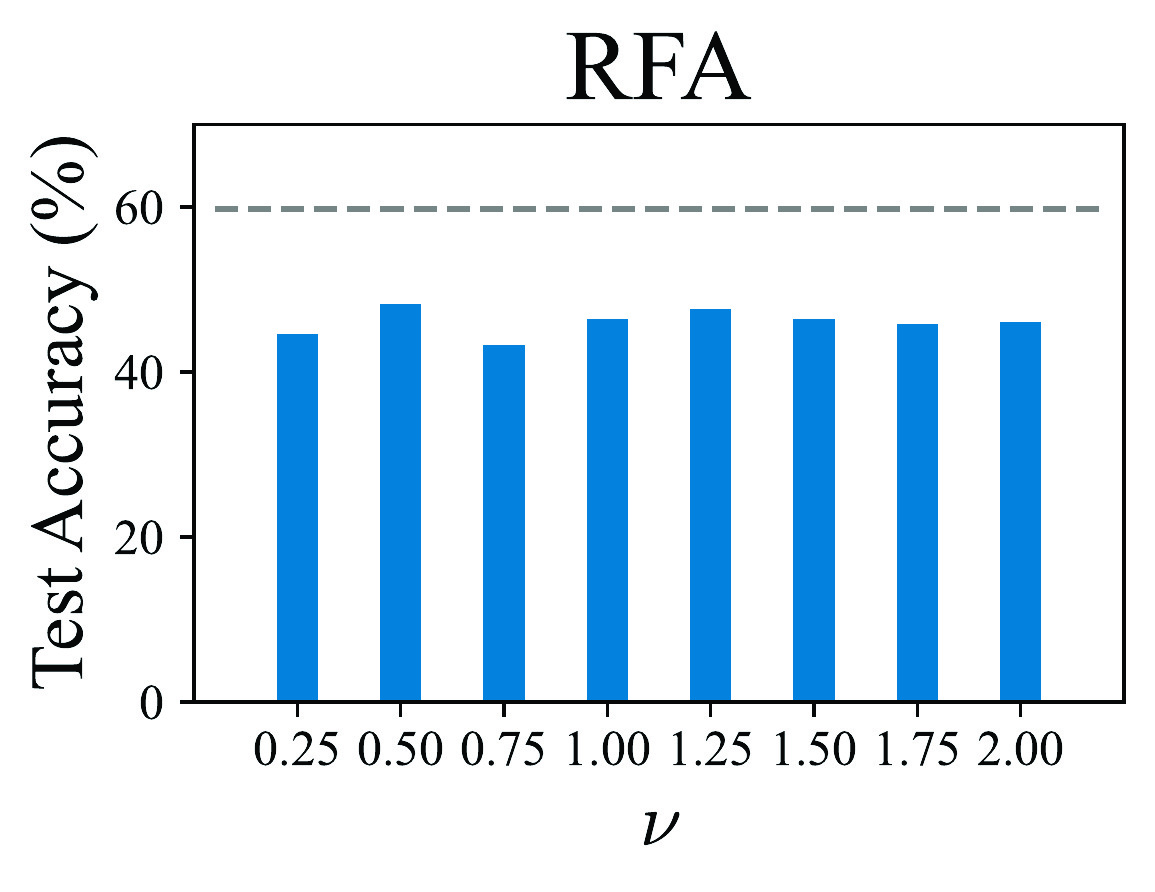}
    \includegraphics[width=.24\linewidth, align=c]{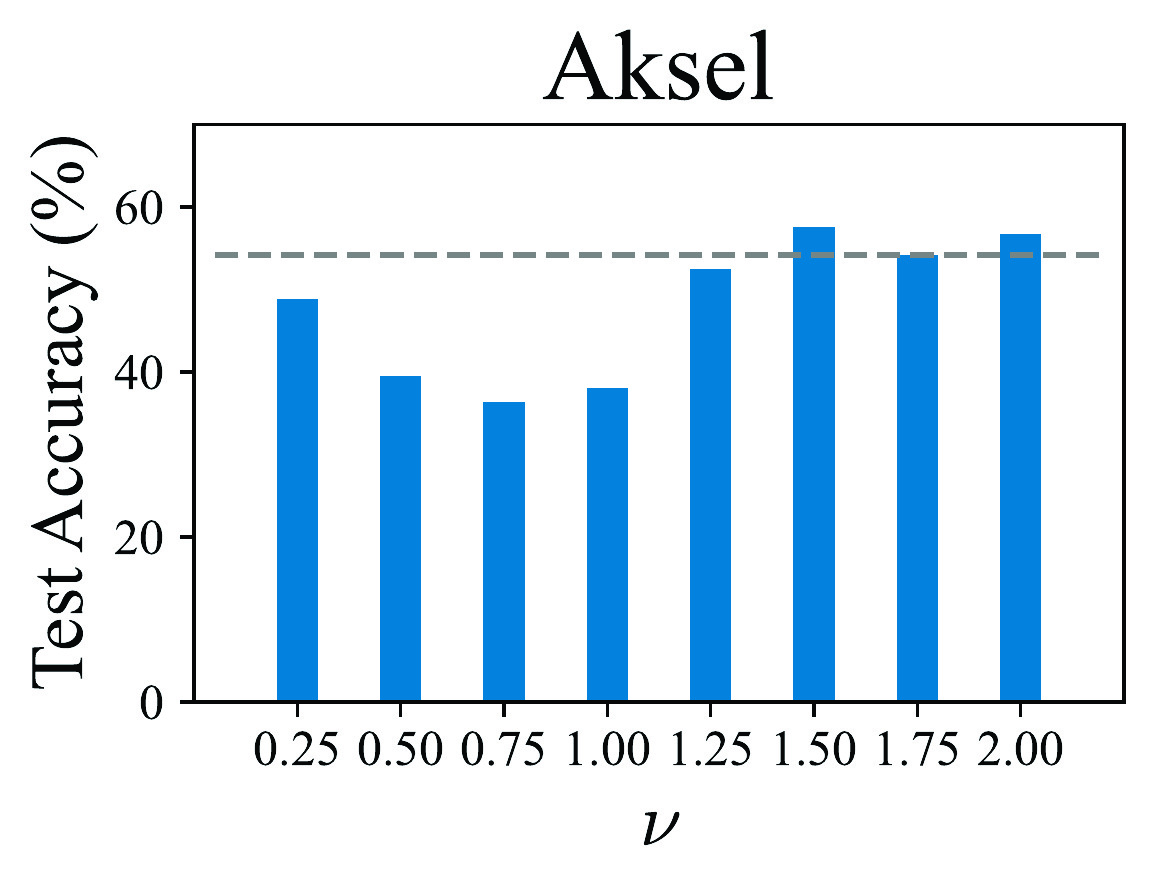}
    \\
    \includegraphics[width=.24\linewidth, align=c]{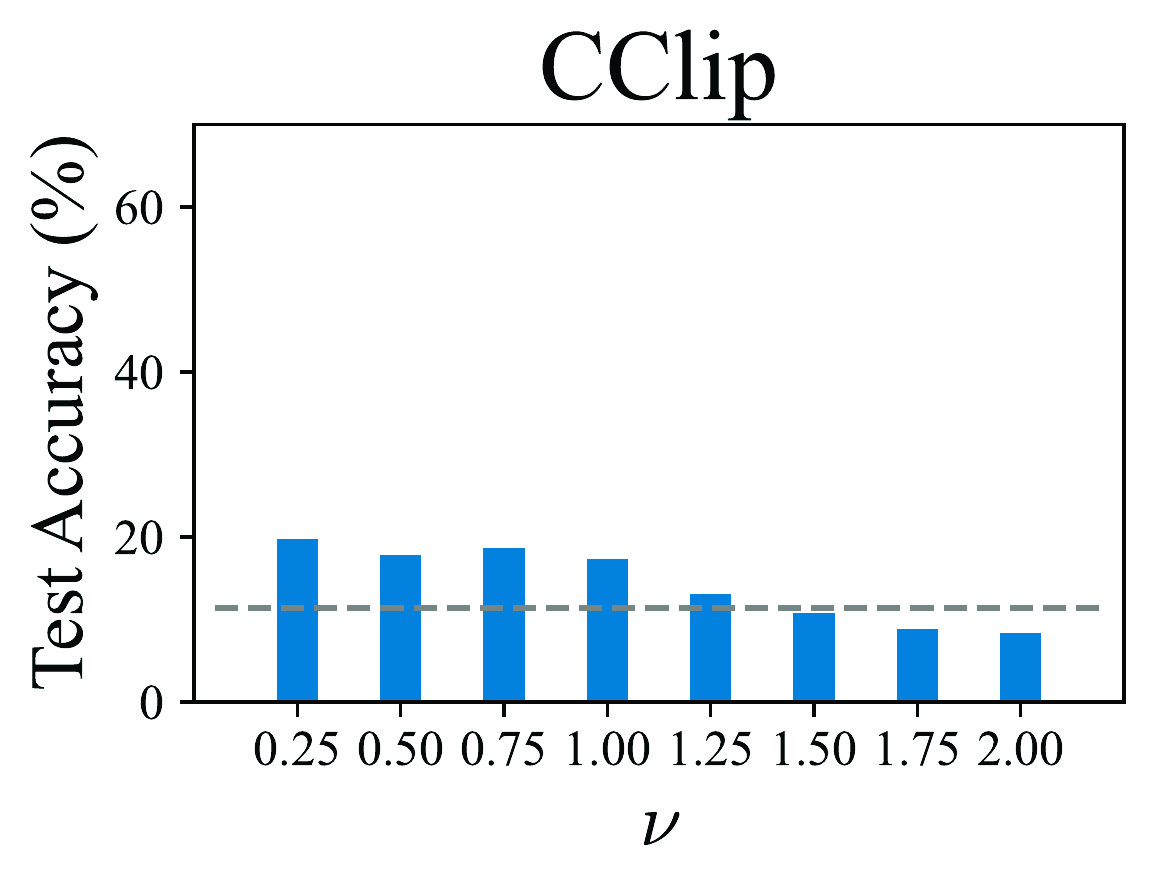}
    \includegraphics[width=.24\linewidth, align=c]{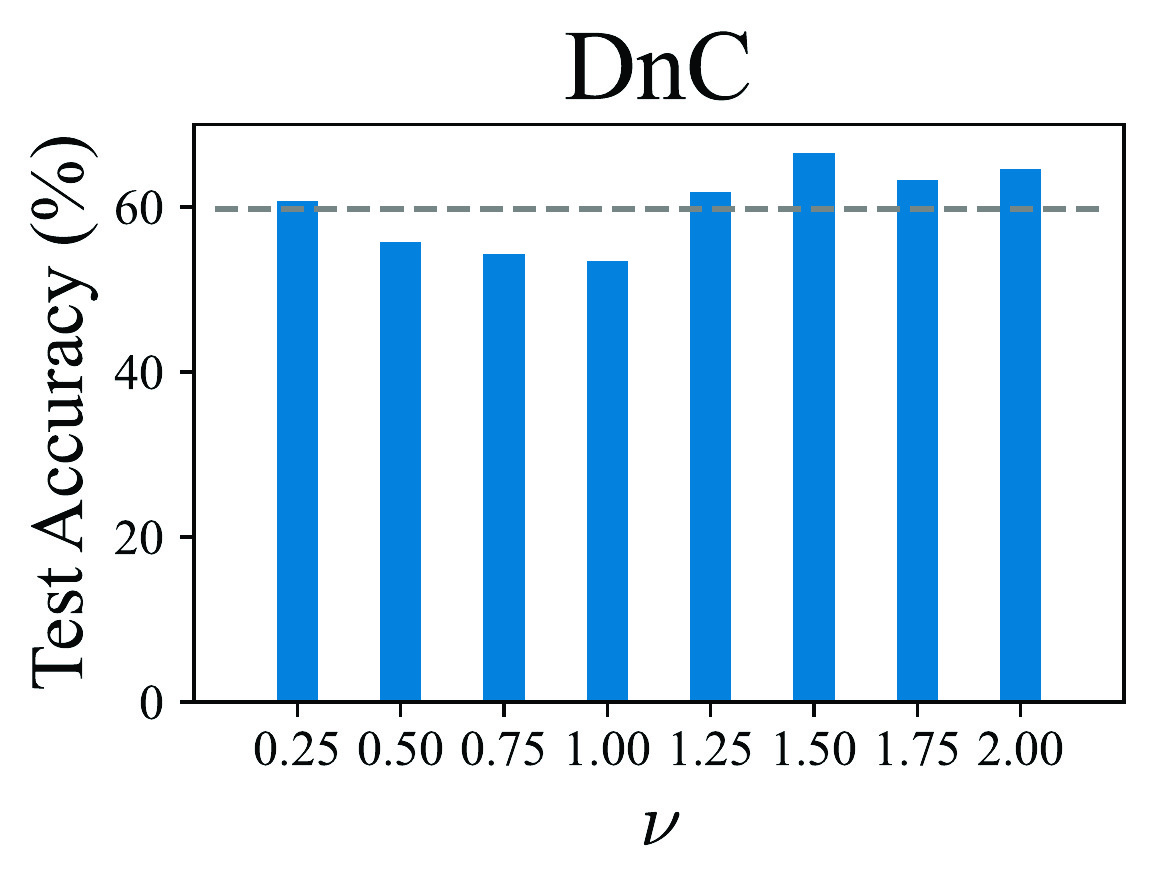}
    \includegraphics[width=.24\linewidth, align=c]{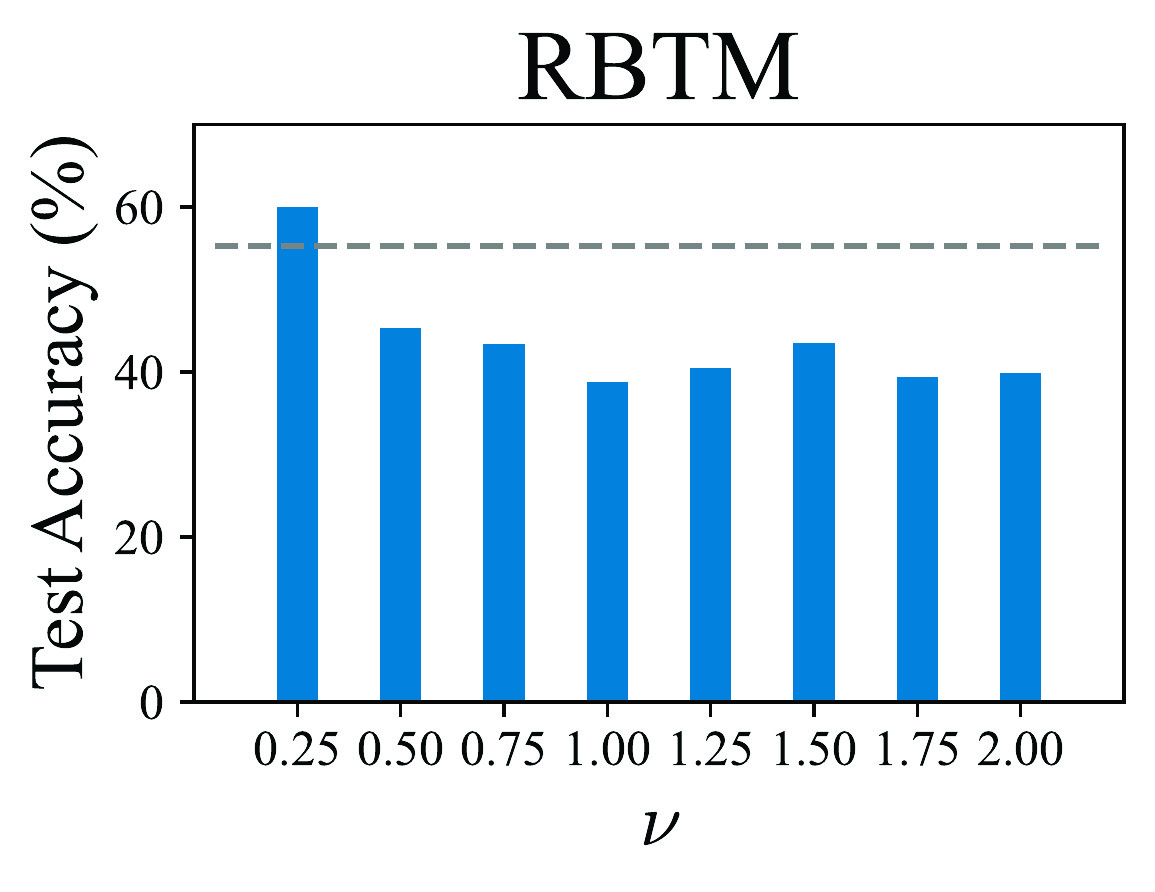}
    \hspace{.05\linewidth}
    \includegraphics[width=.15\linewidth, align=c]{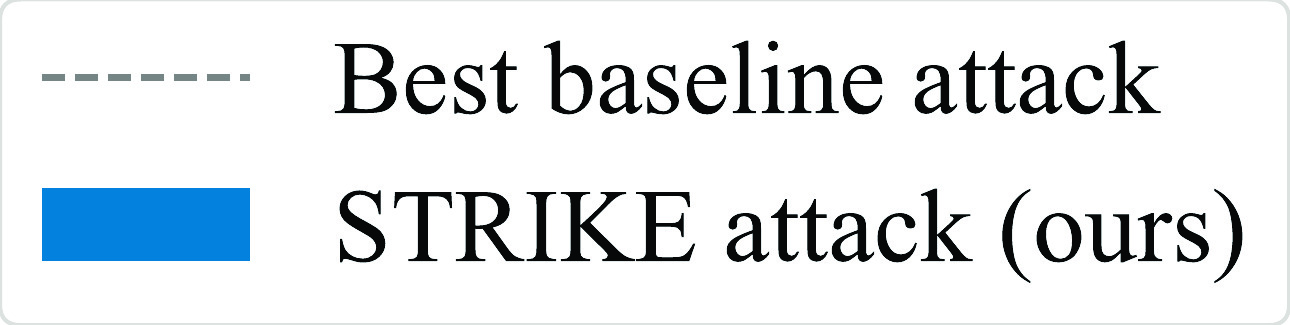}
    
    \caption{
    Accuracy under \SKEW attack with $\nu$ in $\{0.25 * i\mid i=1,\ldots,8\}$ against seven different defenses on ImageNet-12.
    The gray dashed line in each figure represents the lowest test accuracy (best performance) of six baseline attacks introduced in \cref{subsec:experimental_setups}.
    We include it as a reference.
    The \emph{lower} the accuracy, the more effective the attack.
    Other experimental setups align with the main experiment as introduced in \cref{subsec:experimental_setups}.
    }
    \label{fig:nus_results}
\end{figure}

We study the influence of $\nu$ on ImageNet-12 dataset. We report the test accuracy under \SKEW attack with $\nu$ in $\{0.25 * i\mid i=1,\ldots,8\}$ against seven different defenses on ImageNet-12 in \cref{fig:nus_results}.
We also report the lowest test accuracy (best performance) of six baseline attacks introduced in \cref{subsec:experimental_setups} as a reference.
Please note that a \emph{lower} accuracy implies higher attack effectiveness.

As shown in the \cref{fig:nus_results}, the performance of \SKEW is generally competitive with varying $\nu$.
In most cases, simply setting $\nu=1$ can beat other attacks (except for CClip, yet we observe that the performance is low enough to make the model useless).
The impact of $\nu$ value is different for different robust AGRs:
for Median and RFA, the accuracy is relatively stable under different $\nu$s;
for CClip and Multi-Krum, the accuracy is lower with larger $\nu$s;
for Aksel and DnC, the accuracy first decreases and then increases as $\nu$ increases.

\subsubsection{Performance under Different Non-IID Levels}
\label{appsec:niidlevel}
As shown in \cref{tbl:main_experiments}, DnC demonstrates the strongest robustness against various attacks on all datasets.
Therefore, we fix the defense to be DnC in this experiment.
As discussed in \cref{appsec:nus}, simply setting $\nu=1$ yields satisfactory performance of our \SKEW attack.
Thus, we fix $\nu=1$ in this experiment.
We vary Dirichlet concentration parameter $\beta$ in $\set{0.1, 0.2, 0.5, 0.7, 0.9}$ to study how our attack behaves under different non-IID levels.
Lower $\beta$ implies a higher non-IID level.
We additionally test the performance in the IID setting.
Other setups align with the main experiment as introduced in \cref{subsec:experimental_setups}.
The results are posted in \cref{fig:niid_level} below.

As shown in \cref{fig:niid_level}, the accuracy generally increases as $\beta$ decreases for all attacks.
The accuracy under our \SKEW attack is consistently lower than all the baseline attacks.
Besides, we also note that the accuracy gap between our \SKEW attack and other baseline attacks gets smaller when the non-IID level decreases.
We hypothesize the reason is that gradient skew is milder as the non-IID level decreases.
Even in the IID setting, our \SKEW attack is competitive compared to other baselines.

\begin{figure}[H]
    \centering
    \includegraphics[width=.6\linewidth]{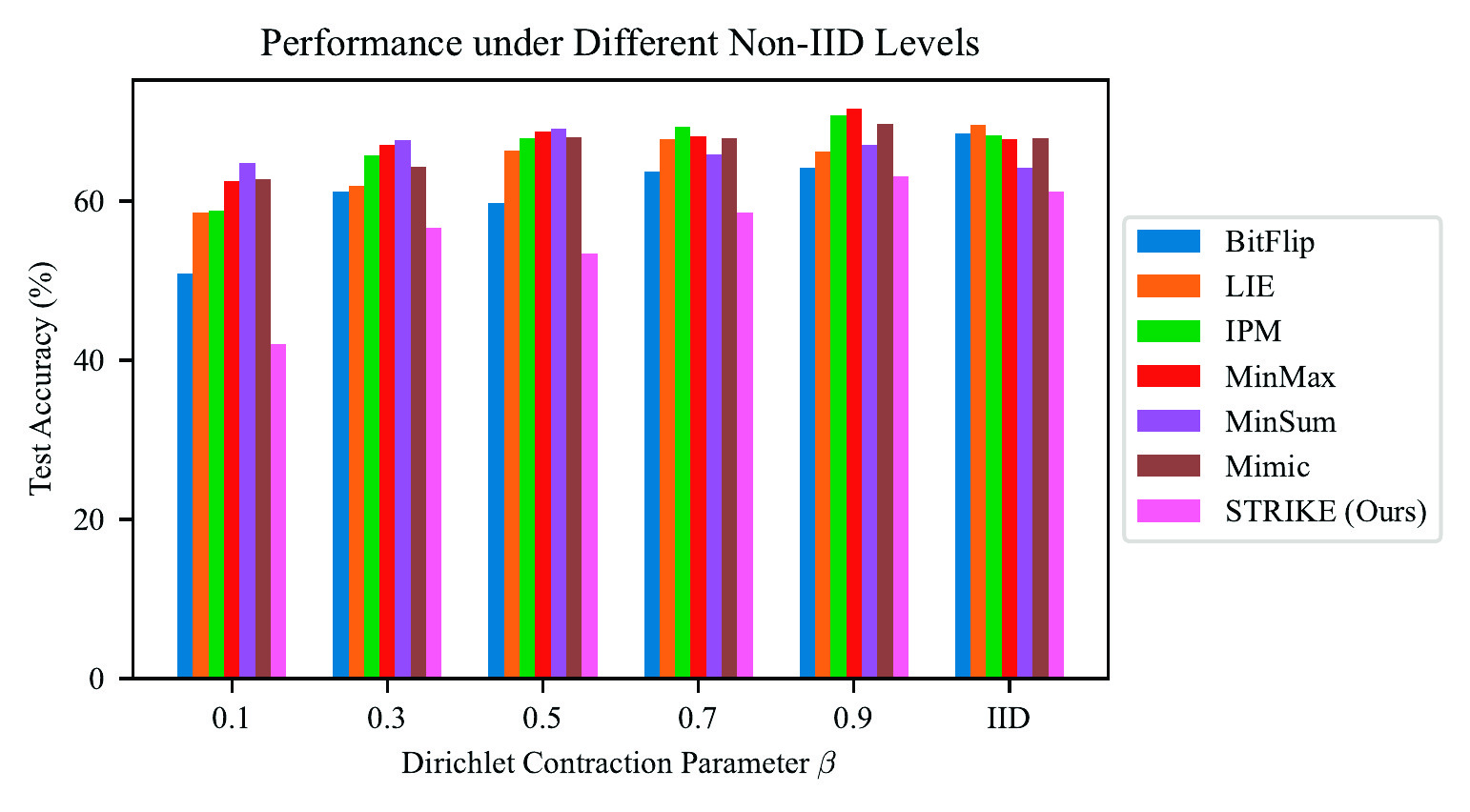}
    \caption{
    Accuracy under different attacks against DnC under different non-IID levels on ImageNet12.
    Lower $\beta$ implies a higher non-IID level.
    "IID" implies that the data is IID distributed.
    The \emph{lower}, the better.
    Other setups align with the main experiment as introduced in \cref{subsec:experimental_setups}.
    }
    \label{fig:niid_level}
\end{figure}

\subsubsection{Performance under Different Byzantine Client Ratio}
\label{appsec:byz_ratio}
As shown in \cref{tbl:main_experiments}, DnC demonstrates the strongest robustness against various attacks on all datasets.
Therefore, we fix the defense to be DnC in this experiment.
As discussed in \cref{appsec:nus}, simply setting $\nu=1$ yields satisfactory performance of our \SKEW attack.
Thus, we fix $\nu=1$ in this experiment.
We vary the number of Byzantine clients $f$ in $\set{5, 10, 15, 20}$ and fix the total number of clients $n$ to be $50$.
In this way, Byzantine client ratio $f/n$ varies in $\set{0.1, 0.2, 0.3, 0.4}$ to study how our attack behaves under different Byzantine client ratio.
Other setups align with the main experiment as introduced in \cref{subsec:experimental_setups}.
The results are posted in \cref{fig:byz_ratio} below.

As shown in \cref{fig:byz_ratio}, the accuracy generally decreases as $f/n$ increases for all attacks.
The accuracy under our \SKEW attack is consistently lower than all the baseline attacks.
The results suggest that all attacks are more effective when there are more Byzantine clients.
Meanwhile, our attack is the most effective under different Byzantine client number.

\begin{figure}[H]
    \centering
    \includegraphics{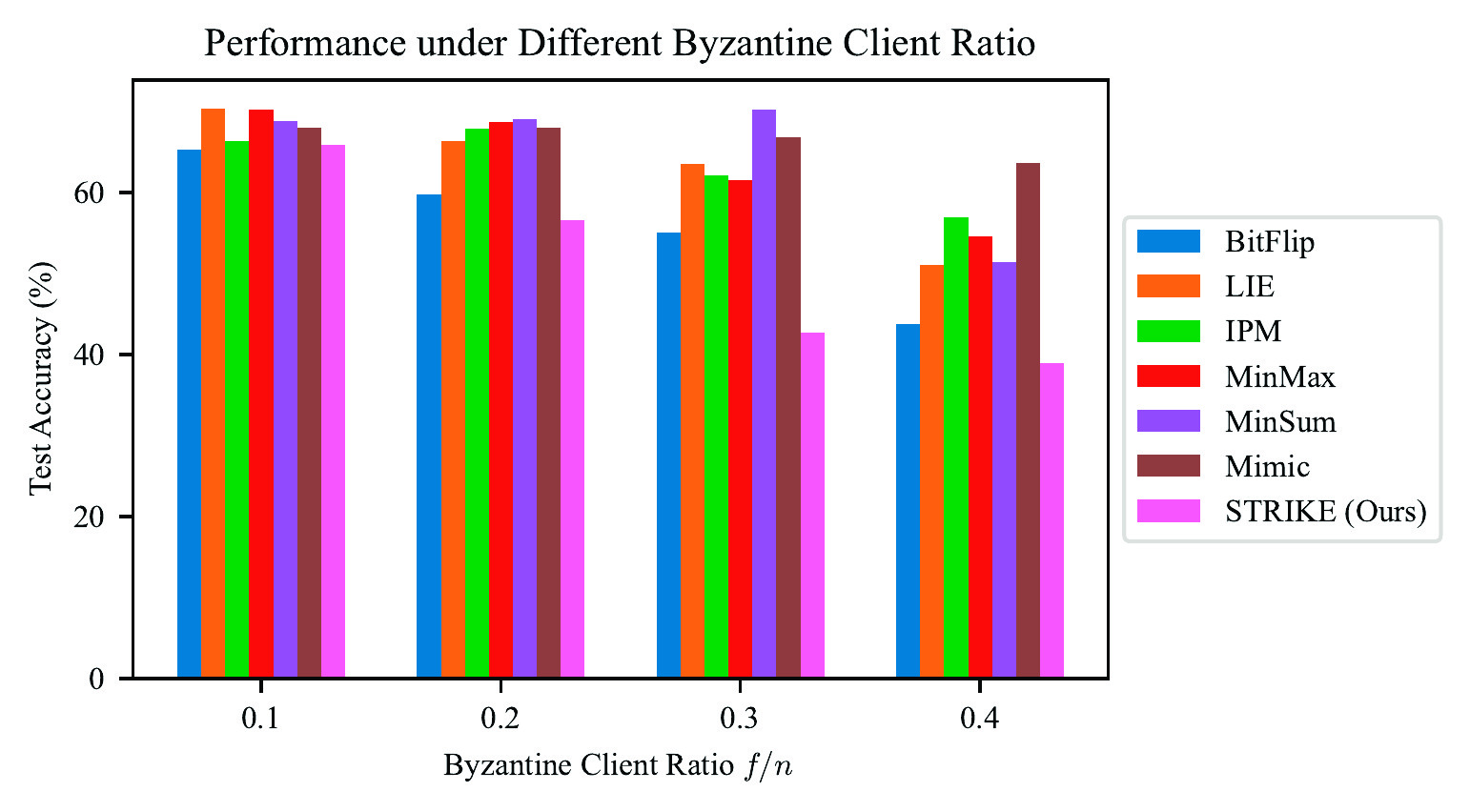}
    \caption{
    Accuracy under different attacks against DnC under different Byzantine client ratio on ImageNet12.
    The \emph{lower}, the better.
    Other setups align with the main experiment as introduced in \cref{subsec:experimental_setups}.
    }
    \label{fig:byz_ratio}
\end{figure}
\subsubsection{Performance under Different Client Number}
\label{appsec:client_num}
As shown in \cref{tbl:main_experiments}, DnC demonstrates the strongest robustness against various attacks on all datasets.
Therefore, we fix the defense to be DnC in this experiment.
As discussed in \cref{appsec:nus}, simply setting $\nu=1$ yields satisfactory performance of our \SKEW attack.
Thus, we fix $\nu=1$ in this experiment.
We vary the number of total clients $n$ in $\set{10, 30, 50, 70, 90}$ and set the number of Byzantine clients $f=0.2n$ accordingly.
In this way, We can study how our attack behaves under different client number.
Other setups align with the main experiment as introduced in \cref{subsec:experimental_setups}.
The results are posted in \cref{fig:client_num} below.

As shown in \cref{fig:client_num}, the accuracy generally decreases as client number $n$ increases for all attacks.
The accuracy under our \SKEW attack is consistently lower than all the baseline attacks under different client number.

\begin{figure}[H]
    \centering
    \includegraphics{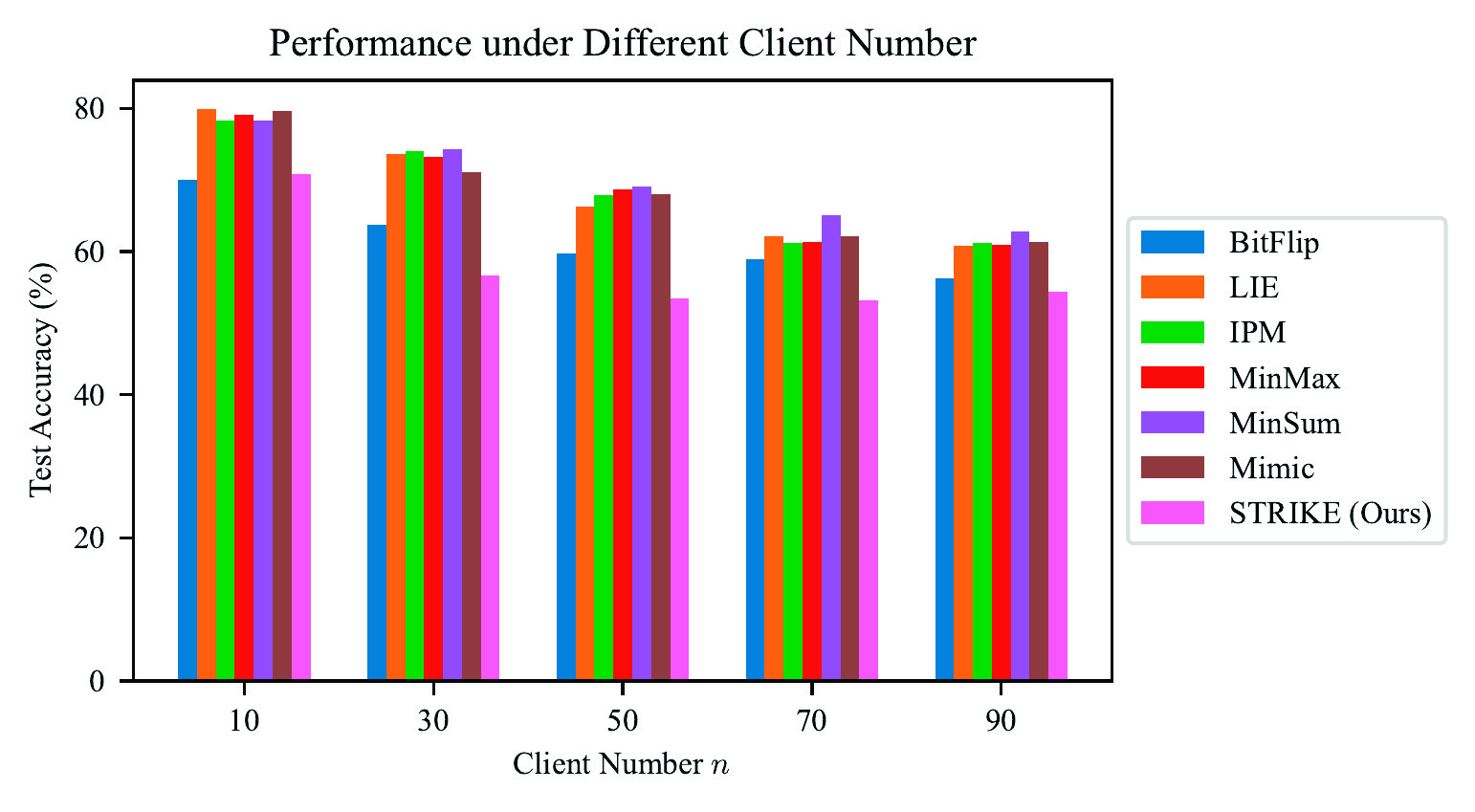}
    \caption{
    Accuracy under different attacks against DnC under different client number on ImageNet12.
    The \emph{lower}, the better.
    Other setups align with the main experiment as introduced in \cref{subsec:experimental_setups}.
    }
    \label{fig:client_num}
\end{figure}

\end{document}